\documentclass{article}

    \usepackage[preprint]{neurips_2020}

\bibpunct{(}{)}{;}{a}{,}{,}

\usepackage[utf8]{inputenc} 
\usepackage[T1]{fontenc}    
\usepackage{hyperref}
\usepackage{url}            
\usepackage{booktabs}       
\usepackage{amsfonts}       
\usepackage{nicefrac}       
\usepackage{microtype}      

\usepackage{algorithm, mathtools}
\usepackage{algorithmic}
\usepackage{bm}
\usepackage{adjustbox}      
\usepackage{threeparttablex} 

\usepackage{todonotes}

\title{Self-Supervised Prototypical Transfer Learning \protect\\ for Few-Shot Classification}

\author{Carlos Medina\thanks{Equal contribution}~~\thanks{Most experiments by CM. Work performed as a semester project at EPFL-INDY supervised by AD, MG.} 
       \And
       Arnout Devos\footnotemark[1] 
       \And
       Matthias Grossglauser \\ \\
       School of Computer and Communication Sciences\\
       \'Ecole Polytechnique F\'ed\'erale de Lausanne (EPFL), Switzerland\\
       \{carlos.medinatemme, arnout.devos, matthias.grossglauser\}@epfl.ch}

\begin{document}

\maketitle

\begin{abstract}
Most approaches in few-shot learning rely on costly annotated data related to the goal task domain during (pre-)training. Recently, unsupervised meta-learning methods have exchanged the annotation requirement for a reduction in few-shot classification performance. Simultaneously, in settings with realistic domain shift, common transfer learning has been shown to outperform supervised meta-learning. Building on these insights and on advances in self-supervised learning, we propose a transfer learning approach which constructs a metric embedding that clusters unlabeled prototypical samples and their augmentations closely together. This pre-trained embedding is a starting point for few-shot classification by summarizing class clusters and fine-tuning. We demonstrate that our self-supervised prototypical transfer learning approach ProtoTransfer outperforms state-of-the-art unsupervised meta-learning methods on few-shot tasks from the mini-ImageNet dataset. In few-shot experiments with domain shift, our approach even has comparable performance to supervised methods, but requires orders of magnitude fewer labels.
\end{abstract}

\section{Introduction}
Few-shot classification \citep{fei2006one} is a learning task in which a classifier must adapt to distinguish novel classes not seen during training, given only a few examples (shots) of these classes. Meta-learning \citep{finn2017model, ren2018meta} is a popular approach for few-shot classification by mimicking the test setting during training through so-called episodes of learning with few examples from the training classes. However, several works \citep{chen2019closer, guo2019new} show that common (non-episodical) transfer learning outperforms meta-learning methods on the realistic cross-domain setting, where training and novel classes come from different distributions.

Nevertheless, most few-shot classification methods still require much annotated data for pre-training.
Recently, several unsupervised meta-learning approaches, constructing episodes via pseudo-labeling \citep{hsu2019unsupervised, ji2019unsupervised} or image augmentations \citep{khodadadeh2019unsupervised, antoniou2019assume, qin2020unsupervised}, have addressed this problem. To our knowledge, unsupervised non-episodical techniques for transfer learning to few-shot tasks have not yet been explored.

Our approach ProtoTransfer performs self-supervised pre-training on an unlabeled training domain and can transfer to few-shot target domain tasks. 
During pre-training, we minimize a pairwise distance loss in order to learn an embedding that clusters noisy transformations of the same image around the original image. Our pre-training loss can be seen as a self-supervised version of the prototypical loss in \cite{snell2017prototypical} in line with contrastive learning, which has driven recent advances in self-supervised representation learning \citep{ye2019unsupervised, chen2020simple, he2019momentum}. 
In the few-shot target task, in line with pre-training, we summarize class information in class prototypes for nearest neighbor inference similar to ProtoNet \citep{snell2017prototypical} and we support fine-tuning to improve performance when multiple examples are available per class.

We highlight our main contributions and results:

\begin{enumerate}
    \item We show that our approach outperforms state-of-the-art unsupervised meta-learning methods by 4\% to 8\% on mini-ImageNet few-shot classification tasks and has competitive performance on Omniglot.
    \item Compared to the fully supervised setting, our approach achieves competitive performance on mini-ImageNet and multiple datasets from the cross-domain transfer learning CDFSL benchmark, with the benefit of not requiring labels during training.
    \item In an ablation study and cross-domain experiments we show that using a larger number of equivalent training classes than commonly possible with episodical meta-learning, and parametric fine-tuning are key to obtaining performance matching supervised approaches.
\end{enumerate}
\section{A Self-Supervised Prototypical Transfer Learning Algorithm}\label{sec:ProtoTransfer}
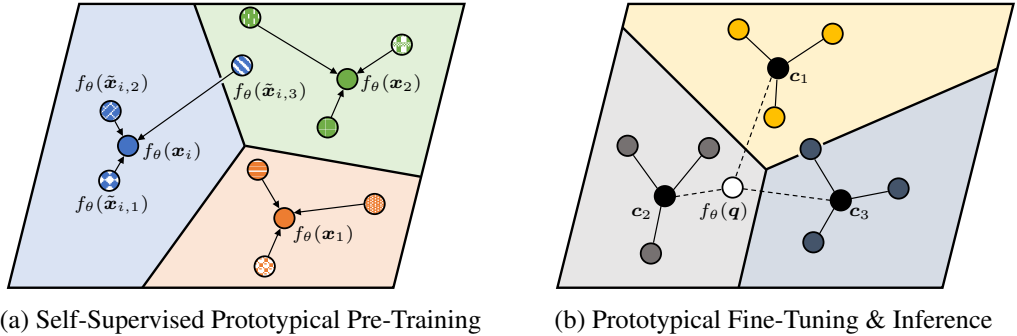
\begin{figure}
\centering
\begin{minipage}{\textwidth}
  \centering
  \setlength{\unitlength}{1cm}
    \begin{picture}(13.5,5)
    \put(0,0){\includegraphics[scale=0.45]{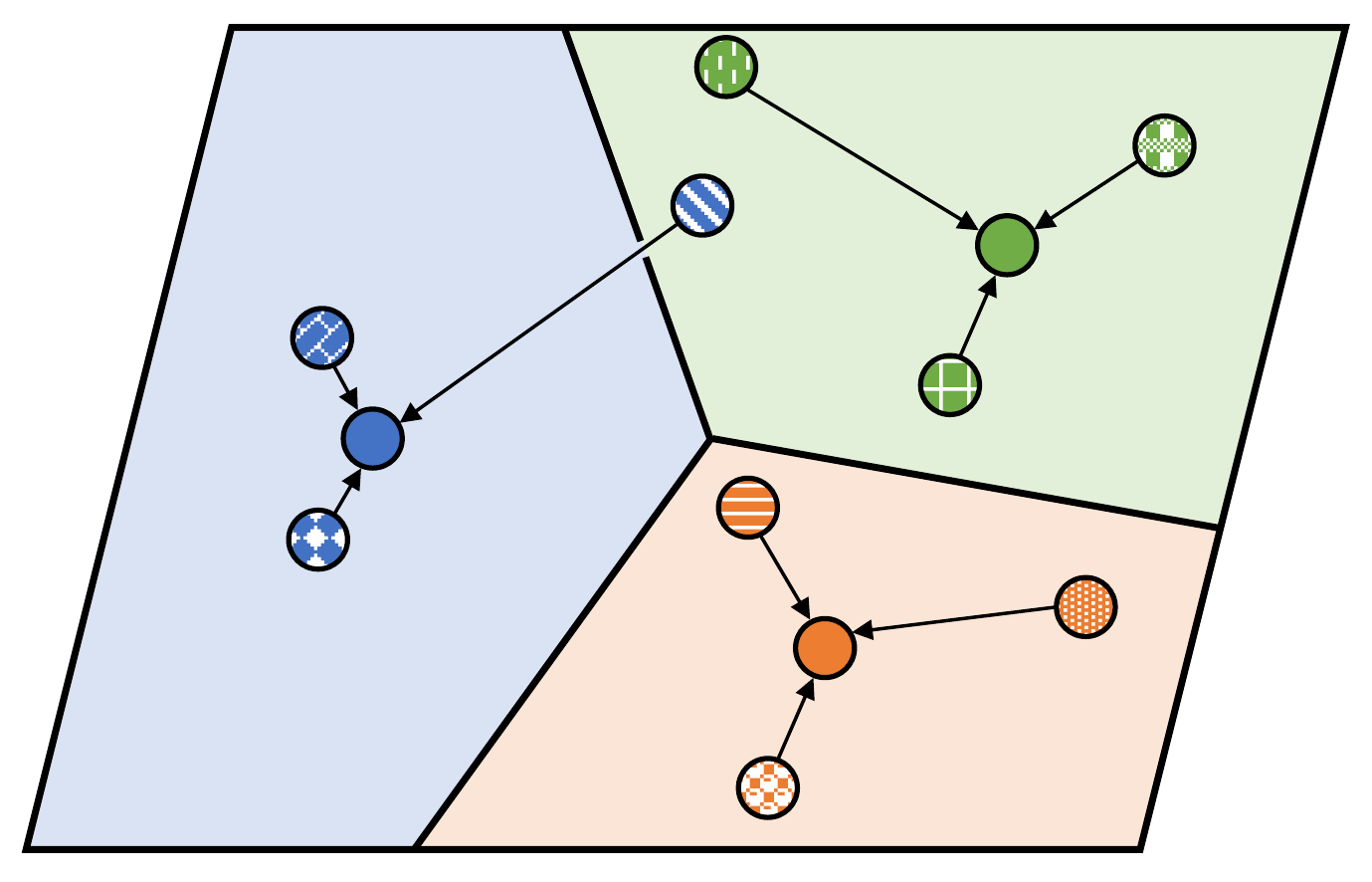}}
    \put(7.3,0){\includegraphics[scale=0.45]{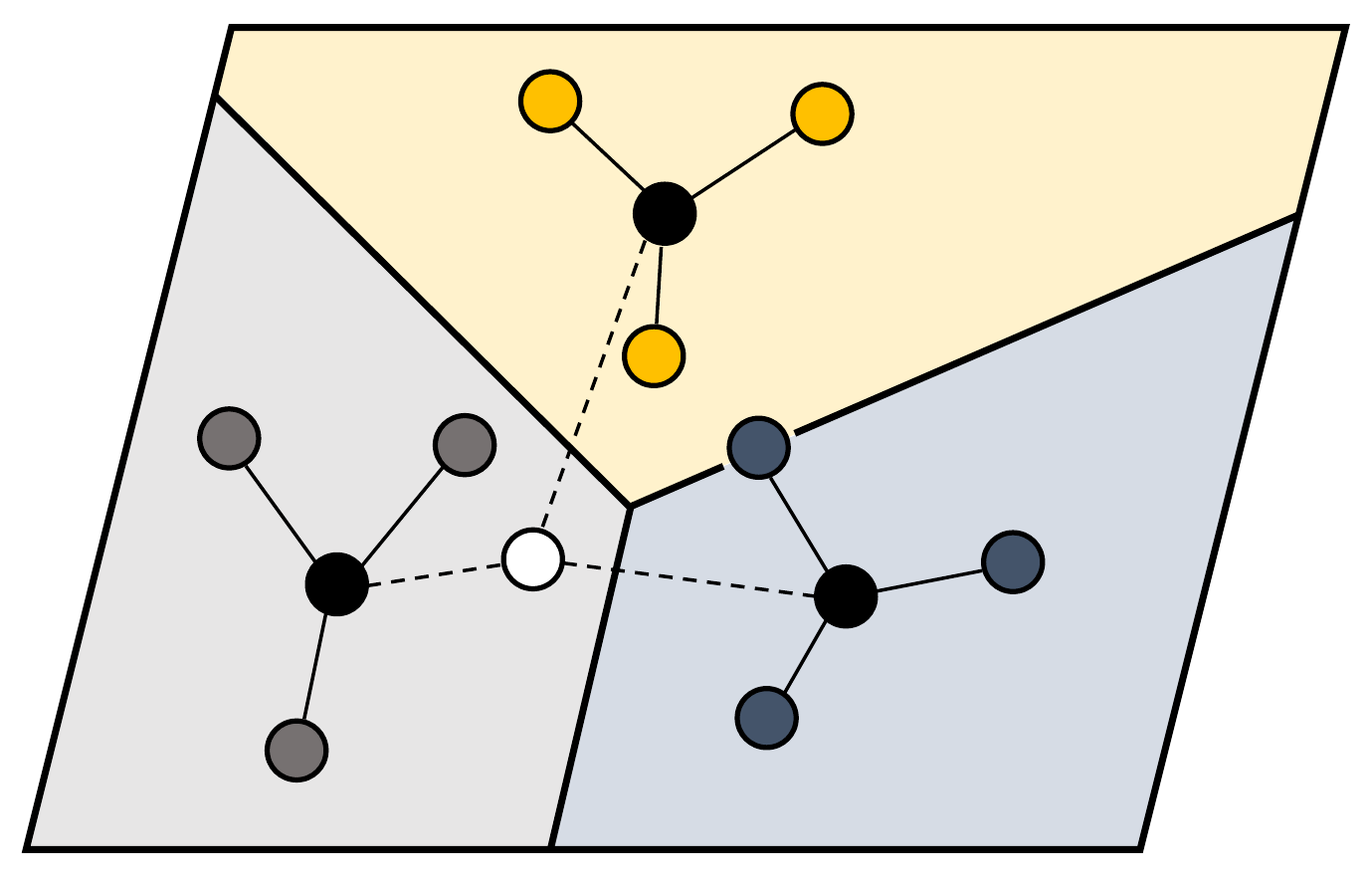}}
    \put(1.9,1.9){$\scriptstyle f_\theta(\bm{x}_i)$}
    \put(1,1.2){$\scriptstyle f_\theta(\tilde{\bm{x}}_{i,1})$}
    \put(1,2.8){$\scriptstyle f_\theta(\tilde{\bm{x}}_{i,2})$}
    \put(3.1,2.7){$\scriptstyle f_\theta(\tilde{\bm{x}}_{i,3})$}
    
    \put(4.8,2.8){$\scriptstyle f_\theta(\bm{x}_{2})$}
    
    \put(3.9,.8){$\scriptstyle f_\theta(\bm{x}_{1})$}
    \put(10.5,2.9){$\scriptstyle \bm{c}_{1}$}
    \put(8.4,1.1){$\scriptstyle \bm{c}_{2}$}
    \put(11.3,1.1){$\scriptstyle \bm{c}_{3}$}
    
    \put(9.3,1.1){$\scriptstyle f_\theta(\bm{q})$}
    \end{picture}\\
    \hspace{-.2cm}(a) Self-Supervised Prototypical Pre-Training \hspace{.8cm} (b) Prototypical Fine-Tuning \& Inference
  \label{fig:sub2}
\end{minipage}
\caption{Self-Supervised Prototypical Transfer Learning. (a): In the embedding, original images $\bm{x}_i$ serve as class prototypes around which their $Q$ augmented views $\tilde{\bm{x}}_{i,q}$ should cluster. (b): Prototypes $\bm{c}_{n}$ are the means of embedded support examples for each class $n$ and initialize a final linear layer for fine-tuning.
An embedded query point $\bm{q}$ is classified via a softmax over the fine-tuned linear layer.}
\label{fig:algorithm}
\end{figure}

Section \ref{sec:preliminaries} introduces the few-shot classification setting and relevant terminology. Further, we describe ProtoTransfer's pre-training stage, ProtoCLR, in Section \ref{sec:pre-training} and its fine-tuning stage, ProtoTune, in Section \ref{sec:fine-tuning}. Figure \ref{fig:algorithm} illustrates the procedure.

\subsection{Preliminaries}\label{sec:preliminaries}
The goal of few-shot classification is to predict classes for a set of unlabeled points (the \textit{query set}) given a small set of labeled examples (the \textit{support set}) from the same classes. Few-shot classification approaches commonly consist of two subsequent learning phases, each using its own set of classes.

The first learning phase utilizes samples from $N_b$ base (training) classes contained within a training set $D_b = \{(\bm{x},y)\}\subset I \times Y_b$, where $\bm{x} \in I$ is a sample with label $y$ in label set $Y_b$. An important aspect of our specific unsupervised learning setting is that the first phase has no access to the per-sample label information, the distribution of classes, nor the size of the label set $Y_b$, for pre-training. This first phase serves as a preparation for the actual few-shot learning in the target domain, i.e. the second learning phase. This second supervised learning phase contains $N_n$ novel (testing) classes as $D_n = \{(\bm{x},y)\}\subset I \times Y_n$, where only few examples for each of the classes in $Y_n$ are available. Concretely, an $N_n$-way $K$-shot classification task consists of $K$ labeled examples for each of the $N_n$ novel classes. In the few-shot learning literature a task is also commonly referred to as an episode.

\subsection{Self-Supervised Prototypical Pre-Training: ProtoCLR}\label{sec:pre-training}

Similar to the few-shot target tasks, we frame every ProtoCLR pre-training learning step as an $N$-way $1$-shot classification task optimized by a contrastive loss function as described below. In this, we draw inspiration from recent progress in unsupervised meta-learning \citep{khodadadeh2019unsupervised} and self-supervised visual contrastive learning of representations \citep{chen2020simple, ye2019unsupervised}.

\begin{algorithm}[t]
\caption{\label{alg:pretraining} Self-Supervised Prototypical Pre-Training (ProtoCLR)}
\begin{algorithmic}[1]
    \STATE \textbf{input:} batch size $N$, augmentations size $Q$, embedding function $f_\theta$, set of random transformations $\mathcal{T}$, step size $\alpha$, distance function $d[\cdot, \cdot]$
    \STATE Randomly initialize $\theta$
    \WHILE{not done}
    \STATE Sample minibatch $\{\bm x_i\}_{i=1}^N$\label{lst:line:start_batch}
    \STATE \textbf{for all} $i\in \{1, \ldots, N\}$ \textbf{do}
    		\STATE $~~~~$\textbf{for all} $q\in \{1, \ldots, Q\}$ \textbf{do}
        		\STATE $~~~~~~~~$draw a random transformation $t \!\sim\! \mathcal{T}$
        		\STATE $~~~~~~~~$$\tilde{\bm x}_{i,q} = t(\bm x_i)$
    \STATE $~~~~$\textbf{end for}
    \STATE \textbf{end for} \label{lst:line:end_batch}
    \STATE \textbf{let} $\ell(i,q) = -\log \frac{\exp(-d[f(\tilde{\bm{x}}_{i,q}), f(\bm{x}_i)])}{\sum_{k=1}^{N} \exp(-d[f(\tilde{\bm{x}}_{i,q}), f(\bm{x}_k)])}$ \\\label{lst:line:start_loss}
     \STATE $\mathcal{L} = \frac{1}{NQ} \sum_{i=1}^N \sum_{q=1}^Q \ell(i,q)$ \label{lst:line:end_loss}
    \STATE $\theta \leftarrow \theta - \alpha \nabla_\theta \mathcal{L}$
    \ENDWHILE
    \STATE \textbf{return} embedding function $f_\theta(\cdot)$
\end{algorithmic}
\end{algorithm}

Algorithm \ref{alg:pretraining} details ProtoCLR and it comprises the following parts: 

\begin{itemize}
    \item Batch generation (Algorithm \ref{alg:pretraining} lines~4-10): Each mini-batch contains $N$ random samples $\{\mathbf{x}_i\}_{i=1...N}$ from the training set. As our self-supervised setting does not assume any knowledge about the base class labels $Y_b$, we treat each sample as it's own class. Thus, each sample ${\bm x}_i$ serves as a 1-shot support sample and class prototype. For each prototype ${\bm x}_i$, $Q$ different randomly augmented versions $\tilde{\bm x}_{i,q}$ are used as query samples.
    
    \item Contrastive prototypical loss optimization (Algorithm \ref{alg:pretraining} lines~11-13): The pre-training loss encourages clustering of augmented query samples $\{\tilde{\bm{x}}_{i,q}\}$ around their prototype ${\bm{x}}_{i}$ in the embedding space through a distance metric $d[\cdot, \cdot]$.
    The softmax cross-entropy loss over $N$ classes is minimized with respect to the embedding parameters $\theta$ with mini-batch stochastic gradient descent (SGD).
\end{itemize}

Commonly, unsupervised pre-training approaches for few-shot classification \citep{hsu2019unsupervised, khodadadeh2019unsupervised, antoniou2019assume, qin2020unsupervised, ji2019unsupervised} rely on meta-learning. Thus, they are required to create small artificial $N$-way ($K$-shot) tasks identical to the downstream few-shot classification tasks. Our approach does not use meta-learning and can use any batch size $N$. Larger batch sizes have been shown to help self-supervised representation learning \citep{chen2020simple} and supervised pre-training for few-shot classification \citep{snell2017prototypical}. We also find that larger batches yield a significant performance improvement for our approach (see Section \ref{sec:ExpAblation}). To generate the query examples, we use image augmentations similar to \citep{chen2020simple} and adjust them for every dataset. The exact transformations are listed in Appendix \ref{app:augmentations}. Following \citet{snell2017prototypical}, we use the Euclidean distance, but our method is generic and works with any metric.

\subsection{Supervised Prototypical Fine-Tuning: ProtoTune}\label{sec:fine-tuning}
After pre-training the metric embedding $f_{\theta}(\cdot)$, we address the target task of few-shot classification. For this, we extend the prototypical nearest-neighbor classifier ProtoNet \citep{snell2017prototypical} with prototypical fine-tuning of a final classification layer, which we refer to as ProtoTune. First, the class prototypes $\bm{c}_n$ are computed as the mean of the class samples in the support set $S$ of the few-shot task:
$$
    \bm{c}_n = \frac{1}{|S_n|}\sum_{(\bm{x}_i, y_i=n) \in S}f_{\theta}(\bm{x}_i)
    \label{eq:prototypes}.
$$
ProtoNet uses non-parametric nearest-neighbor classification with respect to $\bm{c}_n$ and can be interpreted as a linear classifier applied to a learned representation $f_\theta(\mathbf{x})$. Following the derivation in \citet{snell2017prototypical}, we initialize a final linear layer with weights $\mathbf{W}_{n} = 2\mathbf{c}_n$ and biases $b_n = -||\mathbf{c}_n||^2$.
Then, this final layer is fine-tuned with a softmax cross-entropy loss on samples from $S$, while keeping the embedding function parameters $\theta$ fixed. \citet{triantafillou2020meta} proposed a similar fine-tuning approach with prototypical initialization, but their approach always fine-tunes all model parameters.

\section{Experiments}
We carry out several experiments to benchmark and analyze ProtoTransfer. In Section \ref{sec:ExpOmniMini}, we conduct in-domain classification experiments on the Omniglot \citep{lake2011one} and mini-ImageNet \citep{vinyals2016matching} benchmarks to compare to state-of-the-art unsupervised few-shot learning approaches and methods with supervised pre-training. In Section \ref{sec:ExpCDFSL}, we test our method on a more challenging 
cross-domain few-shot learning benchmark \citep{guo2019new}. Section \ref{sec:ExpAblation} contains an ablation study showing how the different components of ProtoTransfer contribute to its performance. In Section \ref{sec:ExpAblationNImages}, we study how pre-training with varying class diversities affects performance. In Section  \ref{sec:ExpGeneralization}, we give insight in generalization from training classes to novel classes from both unsupervised and supervised perspectives. Experimental details can be found in Appendix \ref{app:experimentaldetails} and code is made available\footnote{Our code and pre-trained models are available at \href{https://www.github.com/indy-lab/ProtoTransfer}{\texttt{https://www.github.com/indy-lab/ProtoTransfer}}}.

\subsection{In-Domain Few-shot Classification: Omniglot and mini-ImageNet}\label{sec:ExpOmniMini}
For our in-domain experiments, where the disjoint training class set and novel class set come from the same distribution, we used the popular few-shot datasets Omniglot \citep{lake2011one} and mini-ImageNet \citep{vinyals2016matching}. For comparability we use the Conv-4 architecture proposed in \citet{vinyals2016matching}. Specifics on the datasets, architecture and optmization can be found in Appendices \ref{app:datasets} and \ref{app:architecture_optimization}. We apply limited hyperparameter tuning, as suggested in \citet{oliver2018realistic}, and use a batch size of $N=50$ and number of query augmentations $Q=3$ for all datasets.

In Table \ref{tab:ResultsOmniglotMini}, we report few-shot accuracies on the mini-ImageNet and Omniglot benchmarks. We compare to unsupervised clustering based methods CACTUs \citep{hsu2019unsupervised} and UFLST \citep{ji2019unsupervised} as well as the augmentation based methods UMTRA \citep{khodadadeh2019unsupervised}, AAL \citep{antoniou2019assume} and ULDA \citep{qin2020unsupervised}.
More details on how these approaches compare to ours can be found in Section \ref{sec:RelatedWork}.
Pre+Linear represents classical supervised transfer learning, where a deep neural network classifier is (pre)trained on the training classes and then only the last \textit{linear} layer is fine-tuned on the novel classes.
On mini-ImageNet, ProtoTransfer outperforms all other state-of-the-art unsupervised pre-training approaches by at least 4\% up to 8\% and mostly outperforms the supervised meta-learning method MAML \citep{finn2017model}, while requiring orders of magnitude fewer labels ($NK$ vs $38400+NK$). On Omniglot, ProtoTransfer shows competitive performance with most unsupervised meta-learning approaches.

\begin{table}
    \caption{Accuracy (\%) of unsupervised pre-training methods on $N$-way $K$-shot classification tasks on Omniglot and mini-Imagenet on a Conv-4 architecture. For detailed results, see Tables \ref{tab:omni_full} and \ref{tab:mini_full} in the Appendix. Results style: \textbf{best} and \underline{second best}.}
    \centering
    \begin{tabular}{l c c c c|c c c c}
        \toprule
        {\bf Method \quad  (N,K)} & {\bf (5,1)} & {\bf (5,5)} & \bf{(20,1)} & \bf{(20,5)} & {\bf (5,1)} & {\bf (5,5)} & \bf{(5,20)} & \bf{(5,50)}\\ 
        \midrule
        
        &\multicolumn{4}{c}{\bf Omniglot} & \multicolumn{4}{c}{\bf mini-ImageNet} \\
        \midrule
        {\em Training (scratch)} & 52.50 & 74.78 & 24.91 & 47.62 & 27.59 & 38.48 & 51.53 & 59.63\\
        \midrule
        CACTUs-MAML  & 68.84 &  87.78  & 48.09 & 73.36 & 39.90 & 53.97 & \underline{63.84} & \underline{69.64} \\
        CACTUs-ProtoNet  & 68.12 & 83.58  & 47.75 & 66.27 & 39.18 & 53.36 & 61.54 & 63.55 \\
        UMTRA & 83.80 & 95.43 & \underline{74.25} & \underline{92.12} & 39.93 & 50.73 & 61.11 & 67.15\\
        AAL-ProtoNet & 84.66 & 89.14 & 68.79 & 74.28 &  37.67 & 40.29 & - & - \\
        AAL-MAML++ & \underline{88.40} & \underline{97.96} & 70.21 & 88.32 & 34.57 & 49.18 & - & - \\
        UFLST & \textbf{97.03} & \textbf{99.19} & \textbf{91.28} & \textbf{97.37} & 33.77 & 45.03 & 53.35 & 56.72 \\
        ULDA-ProtoNet & - & - & - & - & 40.63 & \underline{55.41} & 63.16 & 65.20 \\
        ULDA-MetaOptNet & - & - & - & - & \underline{40.71} & 54.49 & 63.58 & 67.65 \\
        ProtoTransfer (ours) & 88.00 & 96.48 & 72.27 & 89.08 & \textbf{45.67} & \textbf{62.99} & \textbf{72.34} & \textbf{77.22} \\
        \midrule
        {\small \em Supervised training} & & & & &  &  & & \\
        {\em MAML} & 94.46 & 98.83 &  84.60 &  96.29 & 46.81 & 62.13 & 71.03 & 75.54\\
        {\em ProtoNet} & 97.70 & 99.28 & 94.40 & 98.39 & 46.44  & 66.33  & 76.73 & 78.91 \\
        {\em Pre+Linear} & 94.30 & 99.08 & 86.05 & 97.11 & 43.87 & 63.01 & 75.46 & 80.17 \\
        
        \bottomrule
        \end{tabular}
    \label{tab:ResultsOmniglotMini}
\end{table}

\subsection{Cross-domain Few-Shot Classification: CDFSL benchmark}\label{sec:ExpCDFSL}
For our cross-domain experiments, where training and novel classes come from different distributions, we turn to the CDFSL benchmark \citep{guo2019new}. This benchmark specifically tests how well methods trained on mini-ImageNet can transfer to few-shot tasks with only limited similarity to mini-ImageNet. In order of decreasing similarity, the four datasets are plant disease images from CropDiseases \citep{mohanty2016using}, satellite images from EuroSAT \citep{helber2019eurosat}, dermatological images from ISIC2018 \citep{tschandl2018ham10000, codella2019skin} and grayscale chest X-ray images from ChestX \citep{wang2017chestx}.
Following \citet{guo2019new}, we use a ResNet-10 neural network architecture. As there is no validation data available for the target tasks in CDFSL, we keep the same ProtoTransfer hyperparameters $N=50$, $Q=3$ as used in the mini-ImageNet experiments.
Experimental details are listed in Appendices \ref{app:cross-domainDatasets} and \ref{app:cross-domainArchOpt}.

For comparison to unsupervised meta-learning, we include our results on UMTRA-ProtoNet and its fine-tuned version UMTRA-ProtoTune \citep{khodadadeh2019unsupervised}. Both use our augmentations instead of those from \citep{khodadadeh2019unsupervised}. For further comparison, we include ProtoNet \citep{snell2017prototypical} for supervised few-shot learning and Pre+Mean-Centroid and Pre+Linear as the best-on-average performing transfer learning approaches from \citet{guo2019new}. As the CDFSL benchmark presents a large domain shift with respect to mini-ImageNet, all model parameters are fine-tuned in ProtoTransfer during the few-shot fine-tuning phase with ProtoTune.

We report results on the CDFSL benchmark in Table \ref{tab:ResultsCDFSL}. ProtoTransfer consistently outperforms its meta-learned counterparts by at least 0.7\% up to 19\% and performs mostly on par with the supervised transfer learning approaches. Comparing the results of UMTRA-ProtoNet and UMTRA-ProtoTune, 
starting from 5 shots, parametric fine-tuning gives improvements ranging from 1\% to 13\%. Notably, on the dataset with the largest domain shift (ChestX), ProtoTransfer outperforms all other approaches.

\begin{table}
    \centering
    \caption{Accuracy (\%) of methods on $N$-way $K$-shot ($N$,$K$) classification tasks of the CDFSL benchmark \citep{guo2019new}. Both our results on methods with unsupervised pre-training (UnSup) and results on methods with supervised pre-training from CDFSL are listed. All models are trained on mini-ImageNet with ResNet-10. For detailed results, see Appendix Table \ref{tab:cdfsl_full}. Results style: \textbf{best} and \underline{second best}.}
    \begin{tabular}{l c c c c |c c c}
        \toprule
        {\bf Method } & \textbf{UnSup} & {\bf (5,5)} & \bf{(5,20)} & \bf{(5,50)} & {\bf (5,5)} & \bf{(5,20)} & \bf{(5,50)}\\ 
        \midrule
        
        & &\multicolumn{3}{c}{\bf ChestX} & \multicolumn{3}{c}{\bf ISIC} \\
        
        \midrule
        ProtoNet &    & 24.05 &28.21  &29.32  & 39.57 & 49.50 &51.99 \\
        Pre+Mean-Centroid & &  \underline{26.31} &30.41 & 34.68 & \underline{47.16} &56.40  & 61.57  \\
        Pre+Linear &  & 25.97 & \underline{31.32} & 35.49 & \textbf{48.11}  & \textbf{59.31} &\textbf{66.48} \\
        UMTRA-ProtoNet& \checkmark  &24.94 & 28.04 & 29.88 & 39.21 & 44.62 & 46.48 \\
        UMTRA-ProtoTune& \checkmark  & 25.00 & 30.41 & \underline{35.63} & 38.47 & 51.60 & 60.12 \\
        ProtoTransfer (ours)& \checkmark  & \textbf{26.71} & \textbf{33.82} & \textbf{39.35}& 45.19 & \underline{59.07} & \underline{66.15}\\
        
        \midrule
        
        & &\multicolumn{3}{c}{\bf EuroSat} & \multicolumn{3}{c}{\bf CropDiseases} \\
        
        \midrule
        
        ProtoNet &   &73.29  &  82.27  &80.48  & 79.72  &88.15  &90.81  \\
        Pre+Mean-Centroid &    & \textbf{82.21} & \underline{87.62} & 88.24 & \underline{87.61} & 93.87 & 94.77 \\
        Pre+Linear &    & \underline{79.08} & \textbf{87.64} & \textbf{91.34} & \textbf{89.25} & \textbf{95.51} & \textbf{97.68} \\
        UMTRA-ProtoNet & \checkmark  & 74.91 & 80.42 & 82.24 & 79.81 & 86.84 & 88.44 \\
        UMTRA-ProtoTune & \checkmark & 68.11 & 81.56 & 85.05 & 82.67 & 92.04 & 95.46 \\
        ProtoTransfer (ours) & \checkmark & 75.62 & 86.80 & \underline{90.46} & 86.53 & \underline{95.06} & \underline{97.01}\\
        \bottomrule
        \end{tabular}
    \label{tab:ResultsCDFSL}
\end{table}

\subsection{Ablation Study: Batch Size, Number of Queries, and Fine-Tuning}\label{sec:ExpAblation}
We conduct an ablation study of ProtoTransfer's components to see how they contribute to its performance.
Starting from ProtoTransfer we successively remove components to arrive at the equivalent UMTRA-ProtoNet which shows similar performance to the original UMTRA approach \citep{khodadadeh2019unsupervised} on mini-ImageNet.
As a reference, we provide results of a ProtoNet classifier on top of a fixed randomly initialized network.

Table \ref{tab:ResultsAblation} shows that increasing the batch size from $N=5$ for UMTRA-ProtoNet to 50 for ProtoCLR-ProtoNet, keeping everything else equal, is crucial to our approach and yields a 5\% to 9\% performance improvement. Importantly, UMTRA-ProtoNet uses our augmentations instead of those from \citep{khodadadeh2019unsupervised}. Thus, this improvement cannot be attributed to using different augmentations than UMTRA.
Increasing the training query number to $Q=3$ gives better gradient information and yields a relatively small but consistent performance improvement.
Fine-tuning in the target domain does not always give a net improvement. Generally, when many shots are available, fine-tuning gives a significant boost in performance as exemplified by ProtoCLR-ProtoTune and UMTRA-MAML in the 50-shot case.
Interestingly, our approach reaches competitive performance in the few-shot regime even before fine-tuning.

\begin{table}
    \centering
    \caption{Accuracy (\%) of methods on $N$-way $K$-shot $(N,K)$ classification tasks on mini-ImageNet with a Conv-4 architecture for different training image batch sizes, number of training queries ($Q$) and optional finetuning on target tasks (FT). UMTRA-MAML results are taken from \cite{khodadadeh2019unsupervised}, where UMTRA uses AutoAugment \citep{cubuk2019autoaugment} augmentations. For detailed results see Table \ref{tab:ResultsAblation_full} in the Appendix. Results style: \textbf{best} and \underline{second best}.}
    \begin{tabular}{l l c c c|c c c c}
        \toprule
        {\bf Training} & {\bf Testing} & {\bf batch size} & {\bf Q} & {\bf FT} & {\bf (5,1)} & {\bf (5,5)} & \bf{(5,20)} & \bf{(5,50)}\\ 
        \midrule
        n.a. & ProtoNet & n.a. & n.a. & no & 27.05 & 34.12 & 39.68 & 41.40 \\
        UMTRA & MAML & $N(=5)$ & 1 & yes & 39.93 & 50.73 & 61.11 & 67.15\\
        UMTRA & ProtoNet & $N(=5)$ & 1 & no & 39.17 & 53.78 & 62.41 & 64.40 \\
        ProtoCLR & ProtoNet & 50 & 1 & no & 44.53 & 62.88 & 70.86 & 73.93 \\
        ProtoCLR & ProtoNet & 50 & 3 & no & 44.89 & \textbf{63.35} & \underline{72.27} & \underline{74.31} \\
        ProtoCLR & ProtoTune & 50 & 3 & yes & \textbf{45.67} & \underline{62.99} & \textbf{72.34} & \textbf{77.22} \\
        \bottomrule
        \end{tabular}
    \label{tab:ResultsAblation}
\end{table}

\subsection{Number of Training Classes and Samples}\label{sec:ExpAblationNImages}
While ProtoTransfer already does not require any labels during pre-training, for some applications, e.g. rare medical conditions, even the collection of sufficiently similar data might be difficult. Thus, we test our approach when reducing the total number of available training images under the controlled setting of mini-ImageNet. Moreover, not all training datasets will have such a diverse set of classes to learn from as the different animals, vehicles and objects in mini-ImageNet. Therefore, we also test the effect of reducing the number of training classes and thereby the class diversity. To contrast the effects of reducing the number of classes or reducing the number of samples, we either remove whole classes from the mini-ImageNet training set or remove the corresponding amount of samples randomly from all classes. The number of samples are decreased in multiples of 600 as each mini-ImageNet class contains exactly 600 samples. We compare the mini-ImageNet few-shot classification accuracies of ProtoTransfer to the popular supervised transfer learning baseline Pre+Linear in Figure \ref{fig:n_images_classes}.

As expected, when uniformily reducing the number of images from all classes (Figure \ref{fig:n_images_classes}a), the few-shot classification accuracy is reduced as well. The performance of ProtoTransfer and the supervised baseline closely match in this case.
When reducing the number of training classes in Figure \ref{fig:n_images_classes}b, ProtoTransfer consistently and significantly outperforms the supervised baseline when the number of mini-ImageNet training classes drops below 16. For example in the 20-shot case with only two training classes, ProtoTransfer outperforms the supervised baseline by a large margin of 16.9\% (64.59\% vs 47.68\%). 
Comparing ProtoTransfer in Figures \ref{fig:n_images_classes}a and \ref{fig:n_images_classes}b, there is only a small difference between reducing images randomly from all classes or taking entire classes away. In contrast, the supervised baseline performance suffers substantially from having fewer classes.

\newcommand\solidrule[1][.5cm]{\rule[0.5ex]{#1}{.4pt}}
\newcommand\dashedrule{\mbox{%
  \solidrule[1mm]\hspace{1mm}\solidrule[1mm]\hspace{1mm}\solidrule[1mm]}}

\begin{figure}
    \centering
    \begin{minipage}{.5\textwidth}
        \centering
        \includegraphics[width=\textwidth]{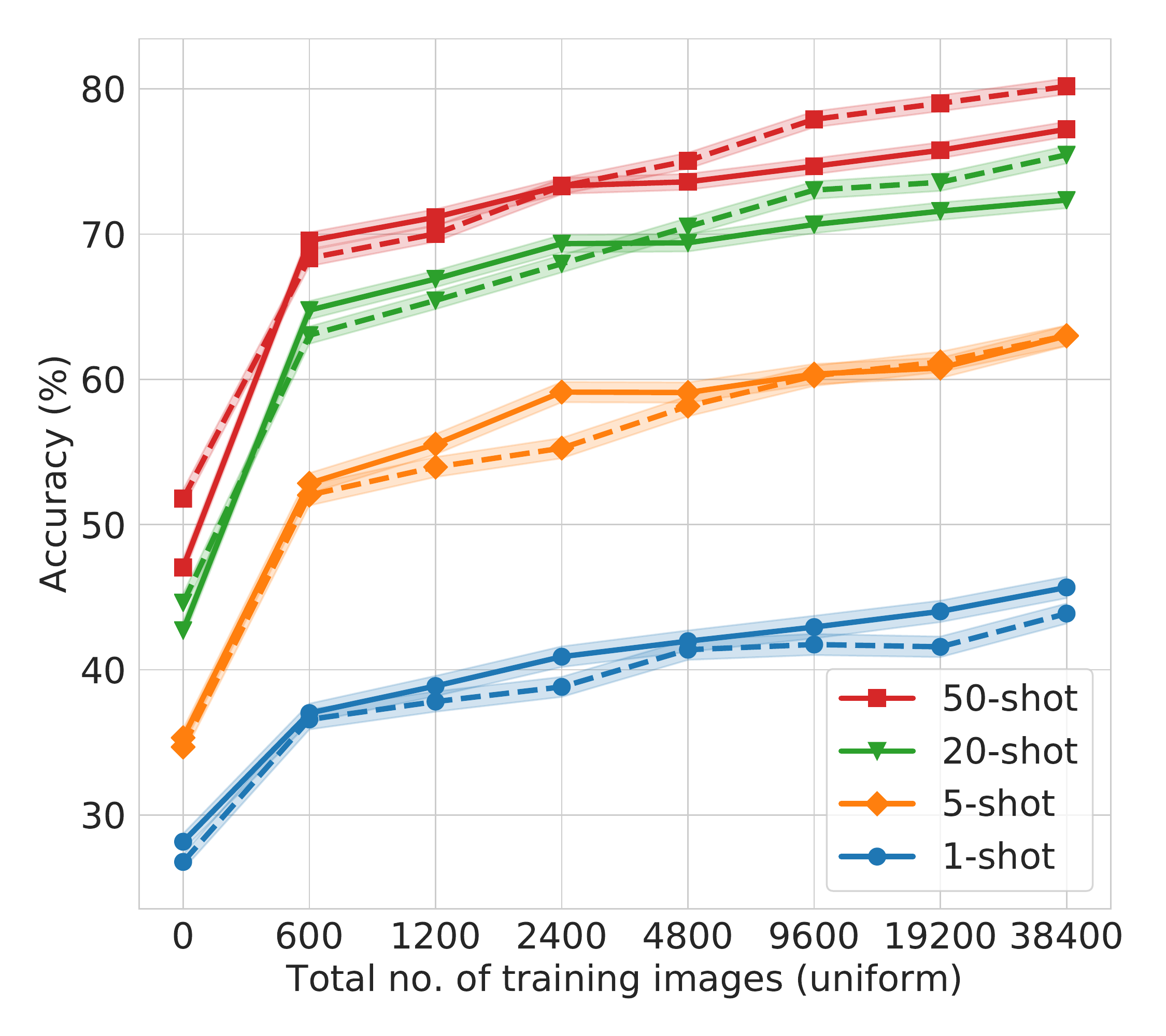}\\
        (a) Varying number of training images.
        \label{fig:n_images}
    \end{minipage}%
    \begin{minipage}{0.5\textwidth}
        \centering
        \includegraphics[width=\textwidth]{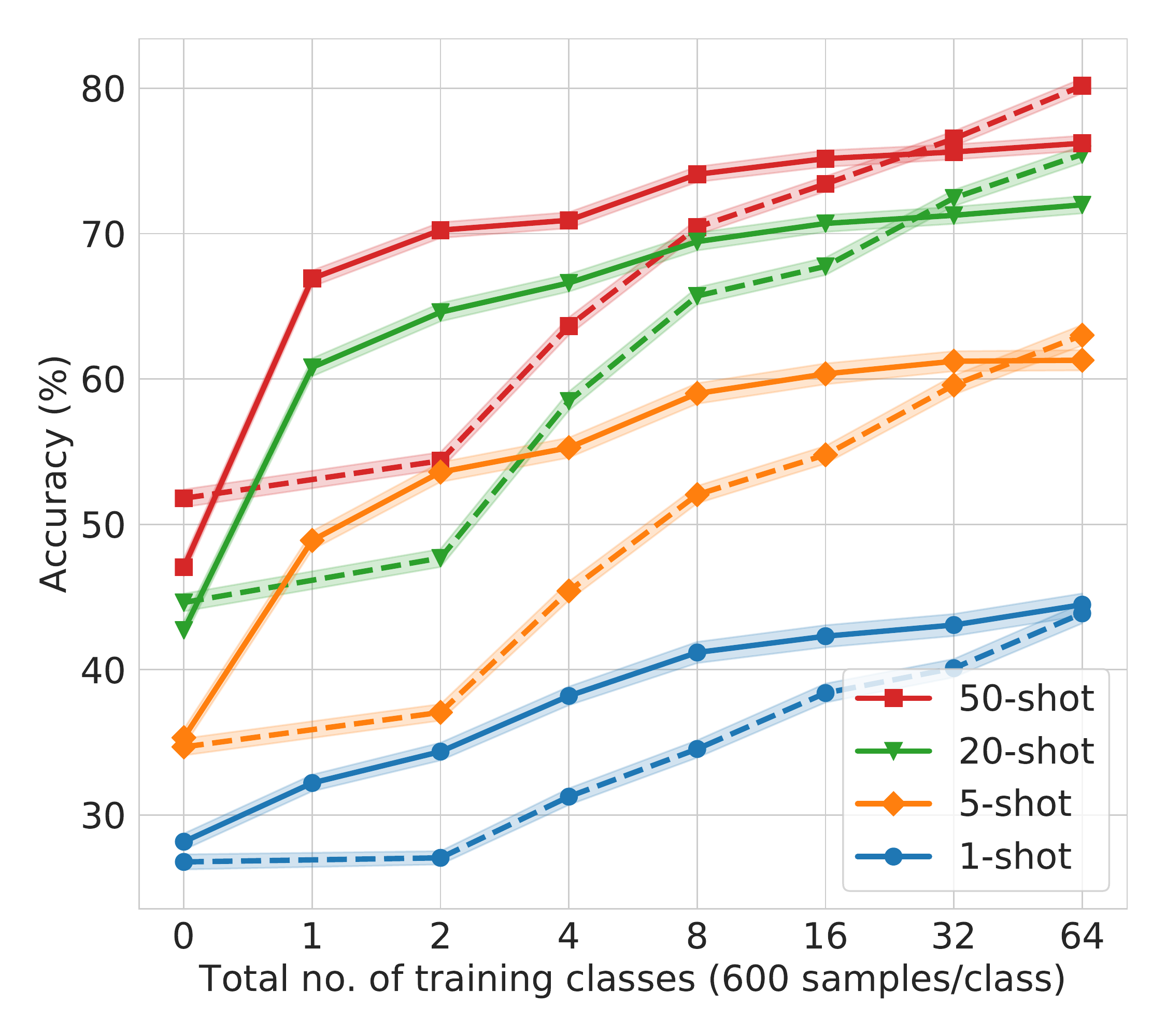}\\
        (b) Varying number of training classes.
        \label{fig:n_classes}
    \end{minipage}
    \caption{$5$-way $K$-shot accuracies with 95\% confidence intervals on mini-ImageNet as a function of training images and classes. Methods:  ProtoTransfer (\solidrule), transfer learning baseline Pre+Linear (\protect\dashedrule). Note the logarithmic scale. Detailed results available in Table \ref{tab:results_n_classes_or_images_full} in the appendix.}
    \label{fig:n_images_classes}
\end{figure}

To validate these in-domain observations in a cross-domain setting, following \citet{devos2019subspace}, we compare few-shot classification performance when training on CUB \citep{welinder2010cub, wah2011caltech} and testing on mini-ImageNet \citep{vinyals2016matching}. CUB consists of 200 classes of birds, while only three of the 64 mini-ImageNet training classes are birds (see \ref{app:cub_data}, \ref{app:cub_opti} for details on CUB). Thus CUB possesses a lower class diversity than 
mini-ImageNet. Table \ref{tab:results_cub2mini} confirms our previous observation numerically and shows that ProtoTransfer has a superior transfer accuracy of 2\% to 4\% over the supervised approach when limited diversity is available in the training classes.

We conjecture that this difference is due to the fact that our self-supervised approach does not make a difference between samples coming from the same or different (latent) classes during training. Thus, we expect it to learn discriminative features despite a low training class diversity. 
In contrast, the supervised case forces multiple images with rich features into the same classes. We thus expect the generalization gap between tasks coming from training classes and testing classes to be smaller with self-supervision.
We provide evidence to support this conjecture in Section \ref{sec:ExpGeneralization}.

\begin{table}
    \centering
    \caption{Accuracy (\%) on $N$-way $K$-shot $(N,K)$ classification tasks on mini-ImageNet for methods trained on the CUB training set (5885 images) with a Conv-4 architecture. All results indicate 95\% confidence intervals over 600 randomly generated test episodes. 
    Results style: \textbf{best} and \underline{second best}.}
    \begin{tabular}{l l |c c c c}
        \toprule
        {\bf Training} & {\bf Testing} & {\bf (5,1)} & {\bf (5,5)} & \bf{(5,20)} & \bf{(5,50)}\\
        \midrule
        ProtoCLR & ProtoNet & \underline{34.56} $\pm$ 0.61 & \textbf{52.76} $\pm$ 0.63 & \underline{62.76} $\pm$ 0.59 & \underline{66.01} $\pm$ 0.55 \\
        ProtoCLR & ProtoTune & \textbf{35.37} $\pm$ 0.63 & \underline{52.38} $\pm$ 0.66 & \textbf{63.82} $\pm$ 0.59 & \textbf{68.95} $\pm$ 0.57 \\
        Pre(training) & Linear & 33.10 $\pm$ 0.60 & 47.01 $\pm$ 0.65 & 59.94 $\pm$ 0.62 & 65.75 $\pm$ 0.63 \\
        \bottomrule
        
        \end{tabular}
    \label{tab:results_cub2mini}
\end{table}

\subsection{Task Generalization Gap}\label{sec:ExpGeneralization}
\begin{figure}
    \centering
    \begin{minipage}{.49\textwidth}
      \centering
      \includegraphics[width=.9\textwidth]{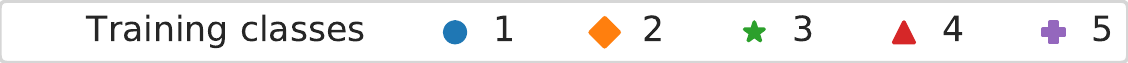}
    \end{minipage}
    \begin{minipage}{.49\textwidth}
      \centering
      \includegraphics[width=.9\textwidth]{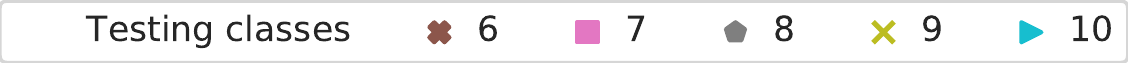}
    \end{minipage}
    \begin{minipage}{.24\textwidth}
        \centering
        \includegraphics[width=\textwidth]{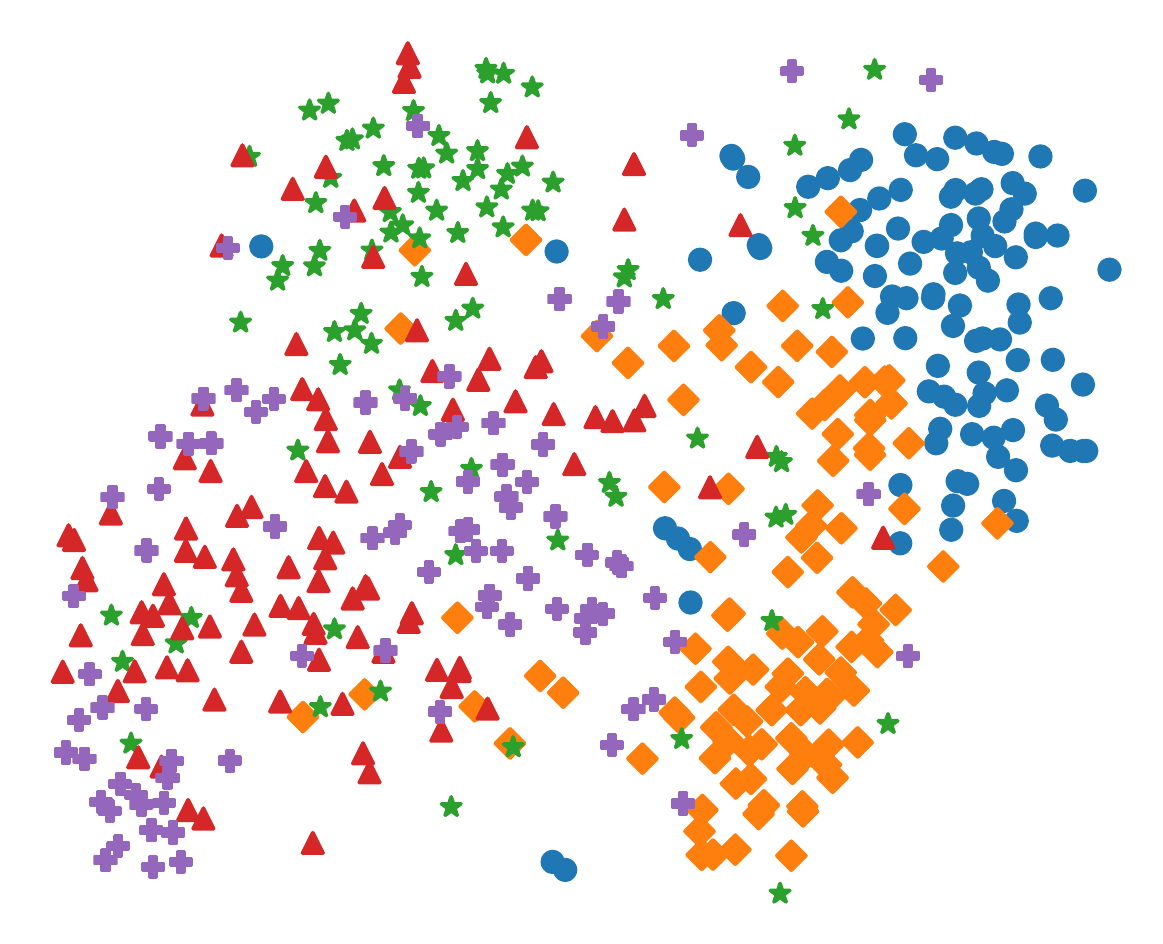}\\
        (a) ProtoCLR training
        \label{fig:ProtoCLRTrainTSNE}
    \end{minipage}%
    \begin{minipage}{0.24\textwidth}
        \centering
        \includegraphics[width=\textwidth]{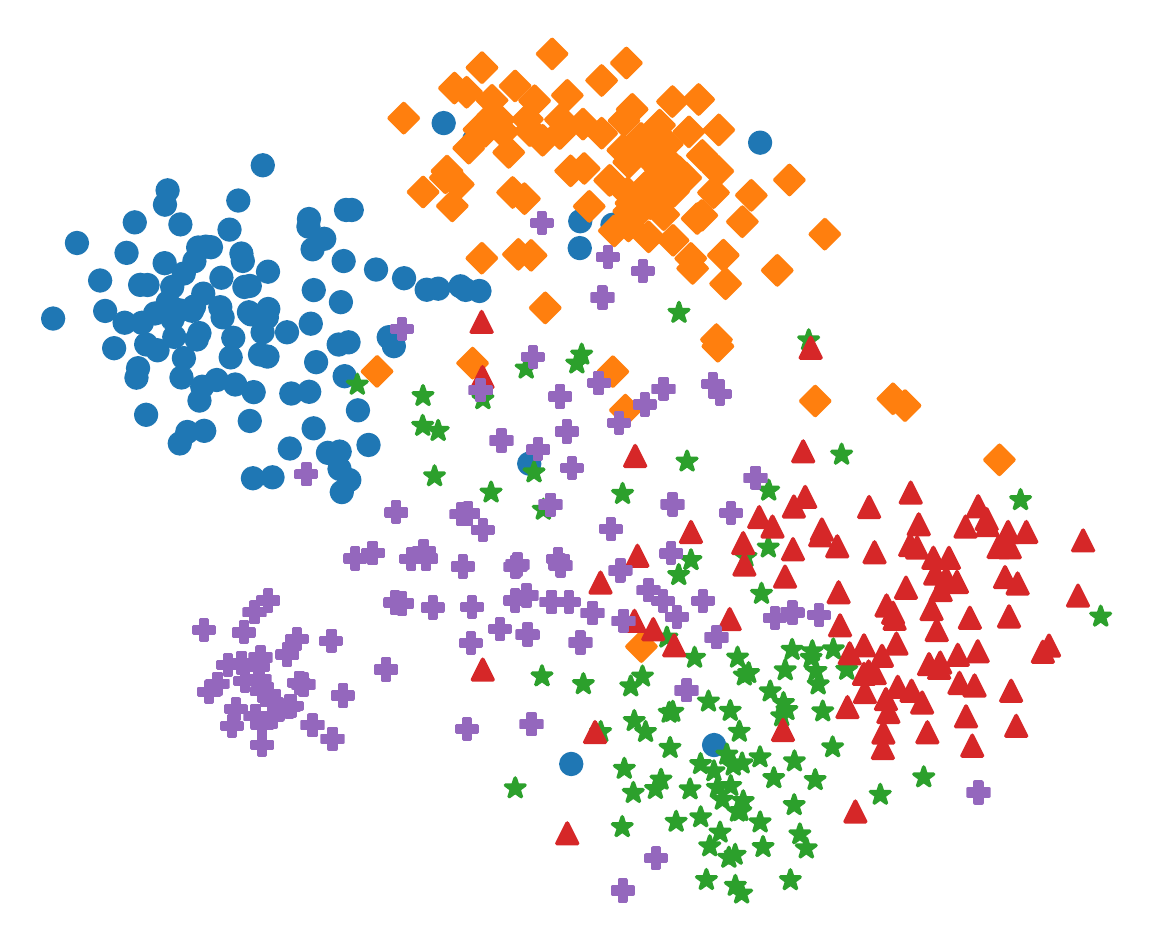}\\
        (b) ProtoNet training
        \label{fig:ProtoNetTrainTSNE}
    \end{minipage}
    \begin{minipage}{0.24\textwidth}
        \centering
        \includegraphics[width=\textwidth]{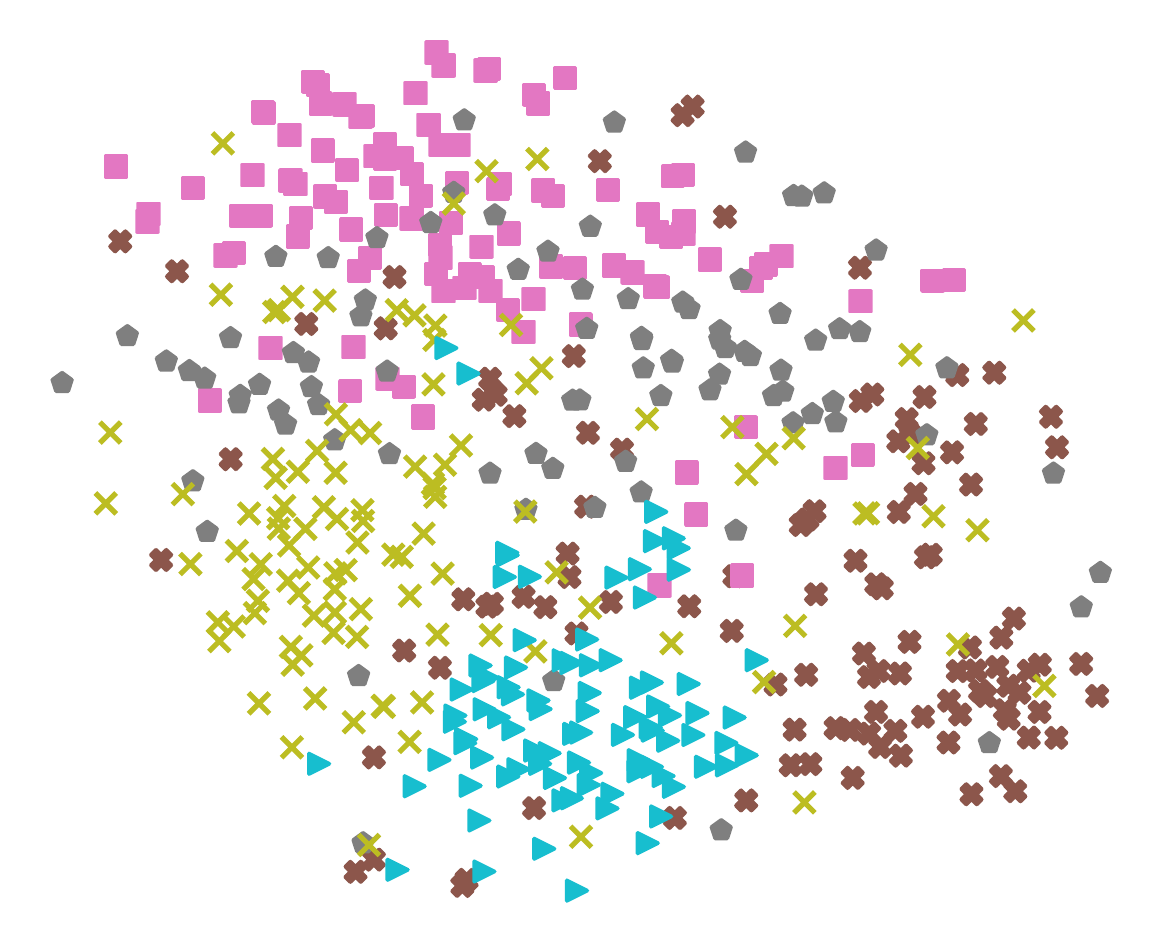}\\
        (c) ProtoCLR testing
        \label{fig:ProtoCLRTestTSNE}
    \end{minipage}
    \begin{minipage}{0.24\textwidth}
        \centering
        \includegraphics[width=\textwidth]{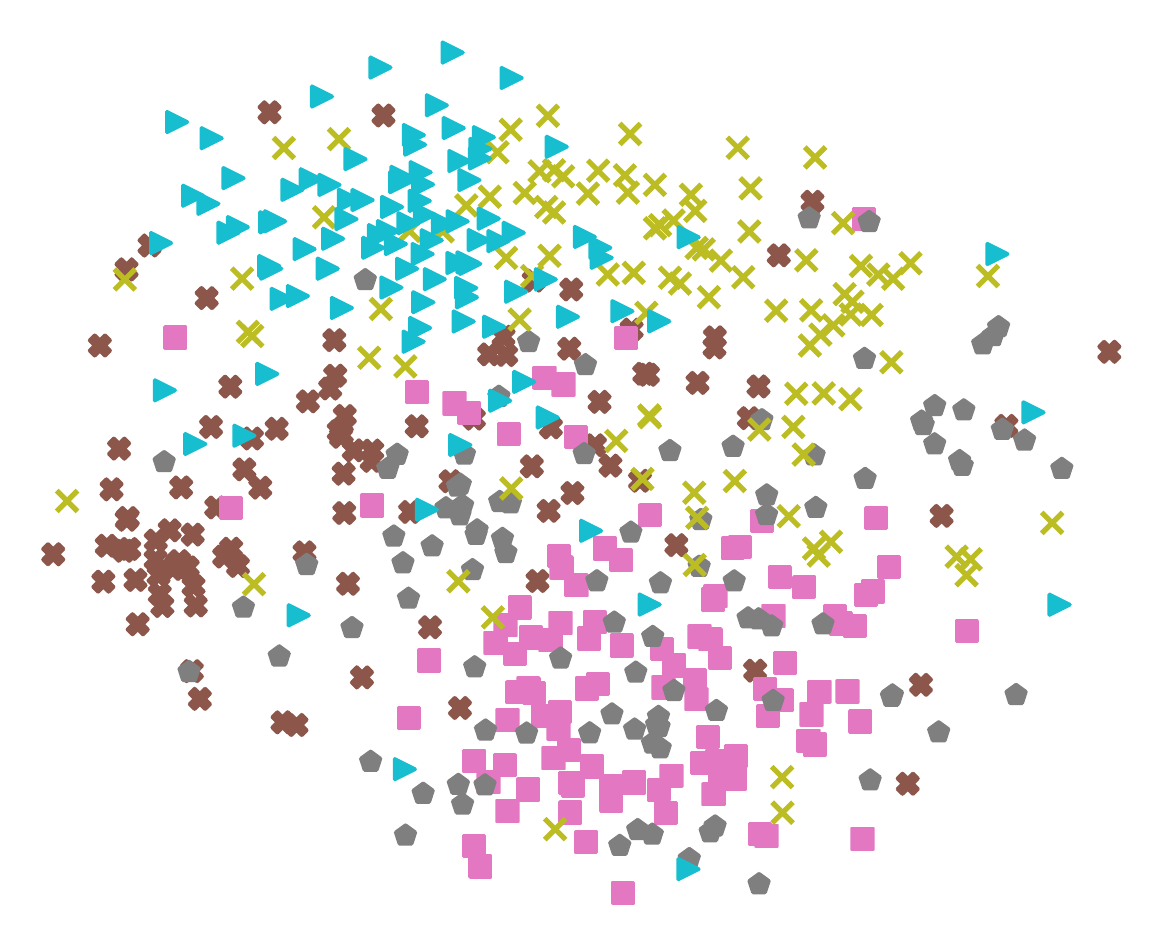}\\
        (d) ProtoNet testing
        \label{fig:ProtoNetTestTSNE}
    \end{minipage}
    \caption{t-SNE plots of trained embeddings on 5 classes from the training and testing sets of mini-ImageNet. Trained embeddings considered are self-supervised ProtoCLR and supervised 20-way 5-shot ProtoNet. For details on the depicted classes, please refer to Appendix \ref{app:t-sne}.}
    \label{fig:taskgeneralizationTSNE}
\end{figure}

To compare the generalization of ProtoCLR with its supervised embedding learning counterpart ProtoNet \citep{snell2017prototypical}, we visualize the learned embedding spaces with t-SNE \citep{maaten2008visualizing} in Figure \ref{fig:taskgeneralizationTSNE}. We compare both methods on samples from 5 random classes from the training and testing sets of mini-ImageNet. In Figures \ref{fig:taskgeneralizationTSNE}a and \ref{fig:taskgeneralizationTSNE}b we observe that, for the same training classes, ProtoNet shows more structure. Comparing all subfigures in Figure \ref{fig:taskgeneralizationTSNE}, ProtoCLR shows more closely related embeddings in Figures \ref{fig:taskgeneralizationTSNE}a and \ref{fig:taskgeneralizationTSNE}c than ProtoNet in Figures \ref{fig:taskgeneralizationTSNE}b and \ref{fig:taskgeneralizationTSNE}d.

These visual observations are supported numerically in Table \ref{tab:ResultsGeneralization}.
Self-supervised embedding approaches, such as UMTRA and our ProtoCLR approach, show a much smaller task generalization gap than supervised ProtoNet.
ProtoCLR shows virtually no classification performance drop.
However, supervised ProtoNet suffers a significant accuracy reduction of 6\% to 12\%.

\begin{table}
    \centering
    \caption{Accuracy (\%) of $N$-way $K$-shot (N,K) classification tasks from the training and testing split of mini-ImageNet. 
    Following \citet{snell2017prototypical}, ProtoNet is trained with $30$-way $1$-shot for $1$-shot tasks and $20$-way $K$-shot otherwise. All results use a Conv-4 architecture. All results show 95\% confidence intervals over 600 randomly generated episodes. }
    \begin{tabular}{l l c|c c c c}
        \toprule
        {\bf Training} & {\bf Testing} & {\bf Data} & {\bf (5,1)} & {\bf (5,5)} & \bf{(5,20)} & \bf{(5,50)}\\ 
        \midrule
        ProtoNet & ProtoNet & Train & 53.74 $\pm$ 0.95 & 79.09 $\pm$ 0.69 & 85.53 $\pm$ 0.53 &  86.62 $\pm$ 0.48\\
        ProtoNet & ProtoNet & Val & 46.62 $\pm$ 0.82 & 67.34 $\pm$ 0.69 & 76.44 $\pm$ 0.57 & 79.00 $\pm$ 0.53 \\
        ProtoNet & ProtoNet & Test & 46.44 $\pm$ 0.78 & 66.33 $\pm$ 0.68 & 76.73 $\pm$ 0.54& 78.91 $\pm$ 0.57\\
        \midrule
        UMTRA & ProtoNet & Train & 41.03 $\pm$ 0.79 & 56.43 $\pm$ 0.78 & 64.48 $\pm$ 0.71 & 66.28 $\pm$ 0.66 \\
        UMTRA & ProtoNet & Test & 38.92 $\pm$ 0.69 & 53.37 $\pm$ 0.68 & 61.69 $\pm$ 0.66 & 65.12 $\pm$ 0.59 \\
        \midrule
        ProtoCLR & ProtoNet & Train & 45.33 $\pm$ 0.63 & 63.47 $\pm$ 0.58 & 71.51 $\pm$ 0.51 & 73.99 $\pm$ 0.49 \\
        ProtoCLR & ProtoNet & Test & 44.89 $\pm$ 0.58 & 63.35 $\pm$ 0.54 & 72.27 $\pm$ 0.45 & 74.31 $\pm$ 0.45 \\
        \bottomrule
        \end{tabular}
    \label{tab:ResultsGeneralization}
\end{table}

\section{Related Work}\label{sec:RelatedWork}
\paragraph{Unsupervised meta-learning:}
Both CACTUs \citep{hsu2019unsupervised} and UFLST \citep{ji2019unsupervised} alternate between clustering for support and query set generation and employing standard meta-learning. In contrast, our method unifies self-supervised clustering and inference in a single model. \citet{khodadadeh2019unsupervised} propose an unsupervised model-agnostic meta-learning approach (UMTRA), where artifical $N$-way $1$-shot tasks are generated by randomly sampling $N$ support examples from the training set and generating $N$ corresponding queries by augmentation. 
\citet{antoniou2019assume} (AAL) generalize this approach to more support shots by randomly grouping augmented images into classes for classification tasks. ULDA \citep{qin2020unsupervised} induce a distribution shift between the support and query set by applying different types of augmentations to each. In contrast, ProtoTransfer uses a single un-augmented support sample, similar to \citet{khodadadeh2019unsupervised}, but extends to several query samples for better gradient signals and steps away from artificial few-shot task sampling by using larger batch sizes, which is key to learning stronger embeddings.

\paragraph{Supervised meta-learning aided by self-supervision:}
Several works have proposed to use a self-supervised loss either alongside supervised meta-learning episodes \citep{gidaris2019boosting, liu2019self} or to initialize a model prior to supervised meta-learning on the source domain \citep{chen2019self, su2019does}. In contrast, we do not require any labels during training.  

\paragraph{Fine-tuning for few-shot classification:} 
\citet{chen2019closer} show that adaptation on the target task is key for good cross-domain few-shot classification performance. Similar to ProtoTune, \citet{triantafillou2020meta} also initialize a final layer with prototypes after supervised meta-learning, but always fine-tune all parameters of the model.

\paragraph{Contrastive loss learning:}
Contrastive losses have fueled recent progress in learning strong embedding functions \citep{ye2019unsupervised, chen2020simple, he2019momentum, tian2020makes, li2020prototypical}. Most similar to our approach is \citet{ye2019unsupervised}. They propose a per-batch contrastive loss that minimizes the distance between an image and an augmented version of it. Different to us, they do not generalize to using multiple augmented query images per prototype and use 2 extra fully connected layers during training.
Concurrently, \citet{li2020prototypical} also use a prototype-based contrastive loss. They compute the prototypes as centroids after clustering augmented images via $k$-Means. They also separate learning and clustering procedures, which ProtoTransfer achieves in a single procedure.

\section{Conclusion}
In this work, we proposed ProtoTransfer for few-shot classification. ProtoTransfer performs transfer learning from an unlabeled source domain to a target domain with only a few labeled examples. Our experiments show that on mini-ImageNet it outperforms all prior unsupervised few-shot learning approaches by a large margin. 
On a more challenging cross-domain few-shot classification benchmark, ProtoTransfer shows similar performance to fully supervised approaches. Our ablation studies show that large batch sizes are crucial to learning good representations for downstream few-shot classification tasks and that parametric fine-tuning on target tasks can significantly boost performance.

\begin{ack}
This work received support from the European Union’s Horizon 2020 research and innovation programme under the Marie Skłodowska-Curie grant agreement No. 754354.
\end{ack}

\vskip 0.2in
\bibliography{neurips}

\newpage

\appendix

\section{Experimental Details}\label{app:experimentaldetails}
\subsection{Datasets} \label{app:datasets}
\subsubsection{In-domain datasets}
For our in-domain experiments we used the popular few-shot datasets Omniglot \citep{lake2011one} and mini-ImageNet \citep{vinyals2016matching}.

Omniglot consists of 1623 handwritten characters from 50 alphabets and 20 examples per character. Identical to \citet{vinyals2016matching}, the grayscale images are resized to 28x28. Following \citet{santoro2016one}, we use 1200 characters for training and 423 for testing. 

Mini-ImageNet is a subset of the ILSVRC-12 dataset \citep{russakovsky2015imagenet}, which contains 60,000 color images that we resized to 84x84. For comparability, we use the splits introduced by \citet{ravi2017optimization} over 100 classes with 600 images each. 64 classes are used for pre-training and 20 for testing. We only use the 16 validation set classes for limited hyperparameter tuning of batch size $N$, number of queries $Q$ and the augmentation strengths.

\subsubsection{Cross-domain datasets}\label{app:cross-domainDatasets}
We evaluate all cross-domain experiments the CDFSL-benchmark \citep{guo2019new}. It comprises four datasets with decreasing similarity to mini-ImageNet. In order of similarity, they are plant disease images from CropDiseases \citep{mohanty2016using}, satellite images from EuroSAT \citep{helber2019eurosat}, dermatological images from ISIC2018 \citep{tschandl2018ham10000, codella2019skin} and grayscale chest x-ray images from ChestX \citep{wang2017chestx}. 

\subsubsection{Caltech-UCSD Birds-200-2011 (CUB) dataset}\label{app:cub_data}
We use the Caltech-UCSD Birds-200-2011 (CUB) dataset \citet{welinder2010cub, wah2011caltech} in our ablation studies. It is composed of 11,788 images from 200 different bird species. We follow the splits proposed by \citet{hilliard2018few} with 100 training, 50 validation and 50 test classes. We do not use the validation set classes.

\subsection{Architecture and Optimization Parameters}\label{app:architecture_optimization}
In the following, we describe the experimental details for the individual experiments. We deliberately stay close to the parameters reported in prior work and do not perform an extensive hyperparameter search for our specific setup, as this can easily lead to performance overestimation compared to simpler approaches \citep{oliver2018realistic}). Table \ref{tab:hyperparameters} summarizes the hyperparameters we used for ProtoTransfer.

\begin{table}
    \centering
    \caption{ProtoTransfer hyperparameter summary.}
    \begin{tabular}{l r r | r r}
        \toprule
        &\multicolumn{2}{c}{in-domain} & \multicolumn{2}{c}{cross-domain} \\
         Hyperparameter & Omniglot & mini-ImageNet & mini-ImageNet & CUB \\
         \midrule
         Model architecture & Conv-4 & Conv-4 & ResNet-10 & Conv-4\\
         Image input size & $28\times28$ & $84\times84$ & $224\times224$ & $84\times84$\\
         Optimizer & Adam & Adam & Adam & Adam\\
         Learning rate & 0.001 & 0.001 & 0.001 & 0.001\\
         Learning rate decay factor  & 0.5 & 0.5 & / & 0.5 \\
         Learning rate decay period  & 25,000 & 25,000 & / & 25,000\\
         Support examples & 1 & 1 & 1 & 1\\
         Augmented queries ($Q$) & 3 & 3 & 3 & 3\\
         Training batch size ($N$)  & 50 & 50 & 50 & 50\\
         Augmentation appendix  & \ref{app:OmniTransforms} & \ref{app:MiniTransforms} & \ref{app:CDFSLTransforms} & \ref{app:MiniTransforms}\\
         \midrule
         Fine-tuning optimizer  & Adam & Adam & Adam & Adam\\
         Fine-tuning learning rate  & 0.001 & 0.001 & 0.001 & 0.001\\
         Fine-tuning batch size  & 5 & 5 & 5 & 5\\
         Fine-tuning epochs  & 15 & 15 & 15 & 15\\
         Fine-tune last layer  & \checkmark & \checkmark & \checkmark & \checkmark \\
         Fine-tune backbone &  &  & \checkmark & \\
        \bottomrule
    \end{tabular}
    \label{tab:hyperparameters}
\end{table}

\subsubsection{In-Domain Experiments}
Our mini-ImageNet and Omniglot experiments use the Conv-4 architecture proposed in \citet{vinyals2016matching} for comparability. Its four convolutional blocks each apply a 64-filters 3x3 convolution, batch normalization, a ReLU nonlinearity and 2x2 max-pooling. The pre-training mostly mirrors \citet{snell2017prototypical} and uses Adam \citep{kingma2015adam} with an initial learning rate of 0.001, which is multiplied by a factor of 0.5 every 25000 iterations. We use a batch size of 50. We do not use the validation set to select the best training epoch. Instead training stops after 20.000 iterations without improvement in training accuracy.

\subsubsection{Cross-Domain Experiments}\label{app:cross-domainArchOpt}
Our experiments on the CDFSL-Challenge are based on the code provided by \citet{guo2019new}. Following \citet{guo2019new}, we use a ResNet10 architecture that is pre-trained on mini-Imagenet images of size 224x224 for 400 epochs with Adam \citep{kingma2015adam} and the default learning rate of 0.001 for best comparability with the results reported in \citet{guo2019new}. The batch size for self-supervised pre-training is 50. We do not use a validation set.

\subsubsection{Caltech-UCSD Birds-200-2011 (CUB) Experiments}\label{app:cub_opti}
The CUB training is identical in terms of architecture (Conv-4) and optimization to the setup for our in-domain experiments.

\subsubsection{Prototypical Fine-Tuning}
During the fine-tuning stage we add a fully connected classification layer after the embedding function and initialize as described in Section \ref{sec:fine-tuning}. We split the support examples into batches of 5 images each and perform 15 fine-tuning epochs with Adam \citep{kingma2015adam} and an initial learning rate of 0.001. For target datasets mini-ImageNet and Omniglot only the last fully connected layer is optimized, while for the CDFSL benchmark experiments the embedding network is adapted as well.

\subsection{Augmentations}\label{app:augmentations}
\subsubsection{CDFSL transforms}\label{app:CDFSLTransforms}
For the CDFSL-benchmark \citep{guo2019new} experiments we employ the same augmentations as \cite{chen2020simple}, as these have proven to work well for ImageNet \citep{russakovsky2015imagenet} images of size 224x224. They are as follows:

\begin{enumerate}
\itemsep0em 
    \item Random crop and resize: \texttt{scale} $\in [0.08, 1.0]$ , \texttt{aspect ratio} $\in [3/4, 4/3]$, Bilinear filter with \texttt{interpolation = 2}
    \item Random horizontal flip
    \item Random ($p=0.8$) color jitter: \texttt{brightness = contrast = saturation = 0.8}, \texttt{hue=0.2}
    \item Random ($p=0.2$) grayscale
    \item Gaussian blur, random radius $\sigma \in [0.1,0.2]$ 
\end{enumerate}

\subsubsection{mini-ImageNet \& CUB transforms}\label{app:MiniTransforms}
For the mini-Imagenet and CUB experiments we used lighter versions of the \citet{chen2020simple} augmentations, namely no Gaussian blur, lower color jitter strengths and smaller rescaling and cropping ranges. They are as follows:

\begin{enumerate}
\itemsep0em 
    \item * Random crop and resize: \texttt{scale} $\in [\bm{0.5}, 1.0]$ , \texttt{aspect ratio} $\in [3/4, 4/3]$, Bilinear filter with \texttt{interpolation = 2}
    \item Random horizontal flip
    \item * Random vertical flip
    \item * Random ($p=0.8$) color jitter: \texttt{brightness = contrast = saturation = 0.4}, \texttt{hue=0.2}
    \item Random ($p=0.2$) grayscale
\end{enumerate}

\subsubsection{Omniglot transforms}\label{app:OmniTransforms}
For Omniglot we use a set of custom augmentations, namely random resizing and cropping, horizontal and vertical flipping, Image-Pixel Dropout \citep{krizhevsky2012imagenet} and Cutout \citep{devries2017improved}.
They are as follows:
\begin{enumerate}
\itemsep0em 
    \item Resize to a size of 28x28 pixels
    \item Random and resize: \texttt{scale} $\in [0.6, 1.0]$ , \texttt{aspect ratio} $\in [3/4, 4/3]$, Bilinear filter with \texttt{interpolation = 2}
    \item Random horizontal flip
    \item Random vertical flip
    \item Random ($p=0.3$) dropout
    \item Random erasing of a rectangular region in an image \citep{zhong2020random}, setting pixel values to \texttt{0}: \texttt{scale} $\in [0.02, 0.33]$, \texttt{aspect ratio} $\in [0.3, 3.3]$
\end{enumerate}

\subsection{Classes for t-SNE Plots}\label{app:t-sne}
The classes in the t-SNE plots are a random subset of classes from the mini-ImageNet base classes (classes 1-5) and the mini-ImageNet novel classes (classes 6-10). Their corresponding labels are the following:
\begin{enumerate}
    \item n02687172 aircraft carrier
    \item n04251144 snorkel
    \item n02823428 beer bottle
    \item n03676483 lipstick
    \item n03400231 frying pan 
    \vspace{.1cm}
    \item n03272010 electric guitar
    \item n07613480 trifle
    \item n03775546 mixing bowl
    \item n03127925 crate
    \item n04146614 school bus
\end{enumerate}

Each of the t-SNE plots in Figure \ref{fig:taskgeneralizationTSNE} shows 500 randomly selected embedded images from within those classes. 

\subsection{Results With Full Confidence Intervals \& References}\label{sec:results_full}

\begin{table}[h]
    \centering
    \caption{Accuracy (\%) of methods on $N$-way $K$-shot classification tasks on Omniglot and a Conv-4 architecture. All results are reported with 95\% confidence intervals over 600 randomly generated test episodes. Results style: \textbf{best} and \underline{second best}.}
    \begin{threeparttable}
    \begin{tabular}{l c c c c}
        \toprule
        {\bf Method \quad  (N,K)} & {\bf (5,1)} & {\bf (5,5)} & \bf{(20,1)} & \bf{(20,5)}\\ 
        \midrule
        
        &\multicolumn{4}{c}{\bf Omniglot} \\
        \midrule
        {\em Training (scratch)} & 52.50  $\pm$ 0.84 & 74.78 $\pm$ 0.69 & 24.91 $\pm$ 0.33 & 47.62 $\pm$ 0.44 \\
        \midrule
        CACTUs-MAML\tnote{\tnote{\footnotemark[1]}}  & 68.84 $\pm$ 0.80 & 87.78 $\pm$ 0.50 & 48.09 $\pm$ 0.41 & 73.36 $\pm$ 0.34 \\
        CACTUs-ProtoNet\tnote{\footnotemark[1]} & 68.12 $\pm$ 0.84 & 83.58 $\pm$ 0.61 & 47.75 $\pm$ 0.43 & 66.27 $\pm$ 0.37 \\
        UMTRA\tnote{\footnotemark[2]} & 83.80 $\pm$ - & 95.43 $\pm$ - & \underline{74.25} $\pm$ - & \underline{92.12} $\pm$ -\\
        AAL-ProtoNet\tnote{\footnotemark[3]} & 84.66 $\pm$ 0.70 & 89.14 $\pm$ 0.27 & 68.79 $\pm$ 1.03 & 74.28 $\pm$ 0.46 \\
        AAL-MAML++\tnote{\footnotemark[3]} & \underline{88.40} $\pm$ 0.75 & \underline{97.96} $\pm$ 0.32 & 70.21 $\pm$ 0.86 & 88.32 $\pm$ 1.22 \\
        UFLST\tnote{\footnotemark[4]} & \textbf{97.03} $\pm$ - & \textbf{99.19} $\pm$ - & \textbf{91.28} $\pm$ - & \textbf{97.37} $\pm$ - \\
        ProtoTransfer (ours) & 88.00 $\pm$ 0.64 & 96.48 $\pm$ 0.26 & 72.27 $\pm$ 0.47 & 89.08 $\pm$ 0.23 \\
        \midrule
        {\small \em Supervised training} & & & & \\
        {\em MAML\tnote{\tnote{\footnotemark[1]}}} & 94.46 $\pm$ 0.35 & 98.83 $\pm$ 0.12 & 84.60 $\pm$ 0.32 & 96.29 $\pm$ 0.13 \\
        {\em ProtoNet} & 97.70 $\pm$ 0.29 & 99.28 $\pm$ 0.10 & 94.40 $\pm$ 0.23 & 98.39 $\pm$ 0.08 \\
        {\em Pre+Linear} & 94.30 $\pm$ 0.43 & 99.08 $\pm$ 0.10 & 86.05 $\pm$ 0.34 & 97.11 $\pm$ 0.11 \\
        \bottomrule
        \end{tabular}
        \begin{tablenotes}\footnotesize
            \item[\tnote{\footnotemark[1]}] \citet{hsu2019unsupervised}
            \item[\tnote{\footnotemark[2]}] \citet{khodadadeh2019unsupervised}
            \item[\tnote{\footnotemark[3]}] \citet{antoniou2019assume}
            \item[\tnote{\footnotemark[4]}] \citet{ji2019unsupervised}
        \end{tablenotes}
        \end{threeparttable}
    \label{tab:omni_full}
\end{table}

\begin{table}[h]
    \centering
    \caption{Accuracy (\%) of methods on $N$-way $K$-shot classification tasks mini-Imagenet and a Conv-4 architecture. All results are reported with 95\% confidence intervals over 600 randomly generated test episodes. Results style: \textbf{best} and \underline{second best}.}
    \begin{threeparttable}
    \begin{tabular}{l c c c c}
        \toprule
        {\bf Method \quad  (N,K)} & {\bf (5,1)} & {\bf (5,5)} & \bf{(5,20)} & \bf{(5,50)}\\ 
        \midrule
        
        & \multicolumn{4}{c}{\bf Mini-ImageNet} \\
        \midrule
        {\em Training (scratch)} & 27.59 $\pm$ 0.59 & 38.48  $\pm$ 0.66 & 51.53 $\pm$ 0.72 & 59.63 $\pm$ 0.74\\
        \midrule
        CACTUs-MAML\tnote{\tnote{\footnotemark[1]}} & 39.90 $\pm$ 0.74 & 53.97 $\pm$ 0.70 & \underline{63.84} $\pm$ 0.70 & \underline{69.64} $\pm$ 0.63 \\
        CACTUs-ProtoNet\tnote{\footnotemark[1]} & 39.18 $\pm$ 0.71 & 53.36 $\pm$ 0.70 & 61.54 $\pm$ 0.68 & 63.55 $\pm$ 0.64 \\
        UMTRA\tnote{\footnotemark[2]} & 39.93 $\pm$ - & 50.73 $\pm$ - & 61.11 $\pm$ - & 67.15 $\pm$ -\\
        AAL-ProtoNet\tnote{\footnotemark[3]} & 37.67 $\pm$ 0.39 & 40.29 $\pm$ 0.68 & - & - \\
        AAL-MAML++\tnote{\footnotemark[3]} & 34.57 $\pm$ 0.74 & 49.18 $\pm$ 0.47 & - & - \\
        UFLST\tnote{\footnotemark[4]} & 33.77 $\pm$ 0.70 & 45.03  $\pm$ 0.73 & 53.35 $\pm$ 0.59 & 56.72 $\pm$ 0.67 \\
        ULDA-ProtoNet\tnote{\footnotemark[5]} & 40.63 $\pm$ 0.61 & \underline{55.41} $\pm$ 0.57 & 63.16 $\pm$ 0.51 & 65.20 $\pm$ 0.50 \\
        ULDA-MetaOptNet\tnote{\footnotemark[5]} & \underline{40.71} $\pm$ 0.62 & 54.49 $\pm$ 0.58 & 63.58 $\pm$ 0.51& 67.65 $\pm$ 0.48 \\
        
        ProtoTransfer (ours) & \textbf{45.67} $\pm$ 0.79 & \textbf{62.99} $\pm$ 0.75 & \textbf{72.34} $\pm$ 0.58 & \textbf{77.22} $\pm$ 0.52 \\
        \midrule
        {\small \em Supervised training} & & & & \\
        {\em MAML\tnote{\tnote{\footnotemark[1]}}} & 46.81 $\pm$ 0.77 & 62.13 $\pm$ 0.72 & 71.03 $\pm$ 0.69 & 75.54 $\pm$ 0.62\\
        {\em ProtoNet} & 46.44 $\pm$ 0.78 & 66.33 $\pm$ 0.68  & 76.73 $\pm$ 0.54 & 78.91 $\pm$ 0.57 \\
        {\em Pre+Linear} & 43.87 $\pm$ 0.69 & 63.01 $\pm$ 0.71 & 75.46 $\pm$ 0.58 & 80.17 $\pm$ 0.51 \\
        \bottomrule
        \end{tabular}
        \begin{tablenotes}\footnotesize
            \item[\tnote{\footnotemark[1]}] \citet{hsu2019unsupervised}
            \item[\tnote{\footnotemark[2]}] \citet{khodadadeh2019unsupervised}
            \item[\tnote{\footnotemark[3]}] \citet{antoniou2019assume}
            \item[\tnote{\footnotemark[4]}] \citet{ji2019unsupervised}
            \item[\tnote{\footnotemark[5]}] \citet{qin2020unsupervised} 
        \end{tablenotes}
        \end{threeparttable}
    \label{tab:mini_full}
\end{table}

\begin{table}[h]
    \centering
    \caption{Accuracy (\%) of methods on $N$-way $K$-shot classification tasks of the CDFSL benchmark \citep{guo2019new}. All models are trained on mini-ImageNet with ResNet-10. All results are reported with 95\% confidence intervals over 600 randomly generated test episodes.  Results style: \textbf{best} and \underline{second best}.}
    \begin{adjustbox}{width=\columnwidth,center}
        \begin{threeparttable}
        \begin{tabular}{l c c c c |c c c}
            \toprule
            {\bf Method} & \textbf{UnSup} & {\bf (5,5)} & \bf{(5,20)} & \bf{(5,50)} & {\bf (5,5)} & \bf{(5,20)} & \bf{(5,50)}\\ 
            \midrule
            
            & \multicolumn{3}{r} {\bf ChestX} \quad & \multicolumn{3}{r}{\bf ISIC} \qquad \\
            
            \midrule
            ProtoNet\tnote{*} & & 24.05 $\pm$ 1.01 & 28.21 $\pm$ 1.15 & 29.32 $\pm$ 1.12 & 39.57 $\pm$ 0.57 & 49.50 $\pm$ 0.55 & 51.99 $\pm$ 0.52 \\
            Pre+Mean-Centroid\tnote{*} & &  \underline{26.31} $\pm$ 0.42 & 30.41 $\pm$ 0.46 & 34.68 $\pm$ 0.46 & \underline{47.16} $\pm$ 0.54 & 56.40 $\pm$ 0.53 & 61.57 $\pm$ 0.66 \\
            Pre+Linear\tnote{*} & & 25.97 $\pm$ 0.41 & \underline{31.32} $\pm$ 0.45 & 35.49 $\pm$ 0.45 & \textbf{48.11} $\pm$ 0.64 & \textbf{59.31} $\pm$ 0.48 &\textbf{66.48} $\pm$ 0.56 \\
            UMTRA-ProtoNet & \checkmark & 24.94 $\pm$ 0.43 & 28.04 $\pm$ 0.44 & 29.88 $\pm$ 0.43 & 39.21 $\pm$ 0.53 & 44.62 $\pm$ 0.49 & 46.48 $\pm$ 0.47 \\
            UMTRA-ProtoTune & \checkmark & 25.00 $\pm$ 0.43 & 30.41 $\pm$ 0.44 & \underline{35.63} $\pm$ 0.48 & 38.47 $\pm$ 0.55 & 51.60 $\pm$ 0.54 & 60.12 $\pm$ 0.50 \\
            ProtoTransfer (ours) & \checkmark & \textbf{26.71} $\pm$ 0.46 & \textbf{33.82} $\pm$ 0.48 & \textbf{39.35} $\pm$ 0.50 & 45.19 $\pm$ 0.56 & \underline{59.07} $\pm$ 0.55 & \underline{66.15} $\pm$ 0.57 \\
            \midrule
            
            &\multicolumn{3}{r}{\bf EuroSat} \quad & \multicolumn{3}{r}{\bf CropDiseases} \\
            
            \midrule
            ProtoNet\tnote{*} & & 73.29 $\pm$ 0.71 & 82.27 $\pm$ 0.57 & 80.48 $\pm$ 0.57 & 79.72 $\pm$ 0.67 & 88.15 $\pm$ 0.51 & 90.81 $\pm$ 0.43 \\
            Pre+Mean-Centroid\tnote{*} & & \textbf{82.21} $\pm$ 0.49 & \underline{87.62} $\pm$ 0.34 & 88.24 $\pm$ 0.29 & \underline{87.61} $\pm$ 0.47 & 93.87 $\pm$ 0.68 & 94.77 $\pm$ 0.34 \\
            Pre+Linear\tnote{*} & & 79.08 $\pm$ 0.61 & \textbf{87.64} $\pm$ 0.47 & \textbf{91.34} $\pm$ 0.37 & \textbf{89.25} $\pm$ 0.51 & \textbf{95.51} $\pm$ 0.31 & \textbf{97.68} $\pm$ 0.21 \\
            UMTRA-ProtoNet & \checkmark & 74.91 $\pm$ 0.72 & 80.42 $\pm$ 0.66 & 82.24 $\pm$ 0.61 & 79.81 $\pm$ 0.65 & 86.84 $\pm$ 0.50 & 88.44 $\pm$ 0.46 \\
            UMTRA-ProtoTune & \checkmark & 68.11 $\pm$ 0.70 & 81.56 $\pm$ 0.54 & 85.05 $\pm$ 0.50 & 82.67 $\pm$ 0.60 & 92.04 $\pm$ 0.43 & 95.46 $\pm$ 0.31 \\
            ProtoTransfer (ours) & \checkmark & 75.62 $\pm$ 0.67 & 86.80 $\pm$ 0.42 & \underline{90.46} $\pm$ 0.37 & 86.53 $\pm$ 0.56 & \underline{95.06} $\pm$ 0.32 & \underline{97.01} $\pm$ 0.26 \\
            \bottomrule
            \end{tabular}
            \begin{tablenotes}\footnotesize
                \item[*] Results from \citet{guo2019new}
            \end{tablenotes}
            \end{threeparttable}
        \end{adjustbox}
    \label{tab:cdfsl_full}
\end{table}

\begin{table}[h]
    \centering
    \caption{Accuracy (\%) of methods on $N$-way $K$-shot classification tasks on Mini-ImageNet with a Conv-4 architecture for different batch sizes, number of training queries ($P$) and optional finetuning on target tasks (FT). UMTRA-MAML uses different augmentations. All results are reported with 95\% confidence intervals over 600 randomly generated test episodes. Results style: \textbf{best} and \underline{second best}.} 
    \begin{adjustbox}{width=\columnwidth,center}
    \begin{threeparttable}
    \begin{tabular}{l l c c c|c c c c}
        \toprule
        {\bf Training} & {\bf Testing} & {\bf batch size} & {\bf P} & {\bf FT} & {\bf (5,1)} & {\bf (5,5)} & \bf{(5,20)} & \bf{(5,50)}\\ 
        \midrule
        UMTRA\tnote{*} & MAML & 5 & 1 & yes & 39.93 $\pm$ - & 50.73 $\pm$ - & 61.11 $\pm$ - & 67.15 $\pm$ - \\
        UMTRA & ProtoNet & 5 & 1 & no & 39.17 $\pm$ 0.53 & 53.78 $\pm$ 0.53 & 62.41 $\pm$ 0.49 & 64.40 $\pm$ 0.46 \\
        ProtoCLR & ProtoNet & 50 & 1 & no & 44.53 $\pm$ 0.60 & 62.88 $\pm$ 0.54 & 70.86 $\pm$ 0.48 & 73.93 $\pm$ 0.44 \\
        ProtoCLR & ProtoNet & 50 & 3 & no & 44.89 $\pm$ 0.58 & \textbf{63.35} $\pm$ 0.54 & \underline{72.27} $\pm$ 0.45 & \underline{74.31} $\pm$ 0.45 \\
        ProtoCLR & ProtoNet & 50 & 5 & no & \underline{45.00} $\pm$ 0.57 & 63.17 $\pm$ 0.55 & 71.70 $\pm$ 0.48 & 73.98 $\pm$ 0.44 \\
        ProtoCLR & ProtoNet & 50 & 10 & no & 44.98 $\pm$ 0.58 & 62.56 $\pm$ 0.53 & 70.78 $\pm$ 0.48 & 73.69 $\pm$ 0.44 \\
        ProtoCLR & ProtoTune & 50 & 3 & yes & \textbf{45.67} $\pm$ 0.76 & \underline{62.99} $\pm$ 0.75 & \textbf{72.34} $\pm$ 0.58 & \textbf{77.22} $\pm$ 0.52 \\
        \bottomrule
        \end{tabular}
        \begin{tablenotes}\footnotesize
            \item[*] \citet{khodadadeh2019unsupervised}
        \end{tablenotes}
        \end{threeparttable}
        \end{adjustbox}
    \label{tab:ResultsAblation_full}
\end{table}

\begin{table}[h]
    \centering
    \caption{Accuracy (\%) of methods on $N$-way $K$-shot classification tasks on Mini-ImageNet with a Conv-4 architecture when reducing the number of pre-training classes or images. All results are reported with 95\% confidence intervals over 600 randomly generated test episodes. Results style: \textbf{best} and \underline{second best}.}
    \begin{adjustbox}{width=\columnwidth,center}
    \begin{threeparttable}
    \begin{tabular}{l c c|c c c c}
        \toprule
        {\bf Method} & {\bf \# images} & {\bf \# classes} & {\bf (5,1)} & {\bf (5,5)} & \bf{(5,20)} & \bf{(5,50)}\\ 
        \midrule
        Random+ProtoTune\tnote{*} & 0 & 0 & 28.16 $\pm$ 0.56 & 35.32 $\pm$ 0.60 & 42.72 $\pm$ 0.63 & 47.05 $\pm$ 0.61 \\
        Random+Linear & 0 & 0 & 26.77 $\pm$ 0.52 & 34.68 $\pm$ 0.58 & 44.62 $\pm$ 0.61 & 51.79 $\pm$ 0.61 \\
        ProtoTransfer & 600 & 1 &  32.20 $\pm$ 0.57 & 48.89 $\pm$ 0.64 & 60.80 $\pm$ 0.62 & 66.91 $\pm$ 0.56 \\
        ProtoTransfer & 600 & $\leq$ 64 & 37.02 $\pm$ 0.65 & 52.84 $\pm$ 0.72 & 64.76 $\pm$ 0.63 & 69.54 $\pm$ 0.58 \\
        Pre+Linear & 600 & $\leq 64$ & 36.58 $\pm$ 0.69 & 52.03 $\pm$ 0.72 & 63.04 $\pm$ 0.68 & 68.34 $\pm$ 0.64 \\
        ProtoTransfer & 1200 & 2 & 34.15 $\pm$ 0.61 & 53.59 $\pm$ 0.68 & 64.59 $\pm$ 0.63 & 70.24 $\pm$ 0.54 \\
        ProtoTransfer & 1200 & $\leq$ 64 & 38.88 $\pm$ 0.70 & 55.53 $\pm$ 0.69 & 66.91 $\pm$ 0.57 & 71.16 $\pm$ 0.56 \\
        Pre+Linear & 1200 & 2 & 27.05 $\pm$ 0.46 & 37.06 $\pm$ 0.57 & 47.68 $\pm$ 0.62 & 54.37 $\pm$ 0.59 \\
        Pre+Linear & 1200 & $\leq 64$ & 37.81 $\pm$ 0.70 & 53.96 $\pm$ 0.69 & 65.43 $\pm$ 0.68 & 70.02 $\pm$ 0.59 \\
        ProtoTransfer & 2400 & 4 & 37.96 $\pm$ 0.64 & 55.27 $\pm$ 0.69 & 66.61 $\pm$ 0.60 & 70.92 $\pm$ 0.55 \\
        ProtoTransfer & 2400 & $\leq$ 64 & 40.90 $\pm$ 0.71 & 59.12 $\pm$ 0.71 & 69.34 $\pm$ 0.60 & 73.32 $\pm$ 0.55 \\
        Pre+Linear & 2400 & 4 & 31.26 $\pm$ 0.57 & 45.41 $\pm$ 0.65 & 58.48 $\pm$ 0.65 & 63.63 $\pm$ 0.61 \\
        Pre+Linear & 2400 & $\leq 64$ & 38.82 $\pm$ 0.69 & 55.26 $\pm$ 0.70 & 67.96 $\pm$ 0.64 & 73.29 $\pm$ 0.58 \\
        ProtoTransfer & 4800 & 8 & 40.74 $\pm$ 0.73 & 59.00 $\pm$ 0.71 & 69.45 $\pm$ 0.61 & 74.08 $\pm$ 0.53 \\
        ProtoTransfer & 4800 & $\leq$ 64 & 41.97 $\pm$ 0.74 & 59.09 $\pm$ 0.71 & 69.40 $\pm$ 0.61 & 73.60 $\pm$ 0.56 \\
        Pre+Linear & 4800 & 8 & 34.54 $\pm$ 0.60 & 52.04 $\pm$ 0.69 & 65.71 $\pm$ 0.59 & 70.44 $\pm$ 0.53 \\
        Pre+Linear & 4800 & $\leq 64$ & 41.38 $\pm$ 0.70 & 58.15 $\pm$ 0.73 & 70.51 $\pm$ 0.63 & 75.05 $\pm$ 0.56 \\
        ProtoTransfer & 9600 & 16 & 42.04 $\pm$ 0.76 & 60.35 $\pm$ 0.72 & 70.70 $\pm$ 0.58 & 75.16 $\pm$ 0.57 \\
        ProtoTransfer & 9600 & $\leq$ 64 & 42.94 $\pm$ 0.78 & 60.36 $\pm$ 0.72 & 70.66 $\pm$ 0.59 & 74.67 $\pm$ 0.55 \\
        Pre+Linear & 9600 & 16 & 38.39 $\pm$ 0.65 & 54.78 $\pm$ 0.67 & 67.75 $\pm$ 0.60 & 73.42 $\pm$ 0.52 \\
        Pre+Linear & 9600 & $\leq 64$ & 41.74 $\pm$ 0.73 & 60.24 $\pm$ 0.68 & 73.03 $\pm$ 0.61 & 77.90 $\pm$ 0.53 \\
        ProtoTransfer & 19200 & 32 & 43.88 $\pm$ 0.76 & 61.22 $\pm$ 0.69 & 71.26 $\pm$ 0.59 & 75.62 $\pm$ 0.52 \\
        ProtoTransfer & 19200 & $\leq$ 64 & 44.02 $\pm$ 0.74 & 60.78 $\pm$ 0.72 & 71.58 $\pm$ 0.56 & 75.77 $\pm$ 0.52 \\
        Pre+Linear & 19200 & 32 & 40.10 $\pm$ 0.63 & 59.58 $\pm$ 0.65 & 72.45 $\pm$ 0.56 & 76.53 $\pm$ 0.52 \\
        Pre+Linear & 19200 & $\leq 64$ & 41.58 $\pm$ 0.71 & 61.20 $\pm$ 0.66 & 73.57 $\pm$ 0.56 & 79.01 $\pm$ 0.51 \\
        ProtoTransfer & 38400 & 64 & 45.67 $\pm$ 0.76 & 62.99 $\pm$ 0.75 & 72.34 $\pm$ 0.58 & 77.22 $\pm$ 0.52 \\
        Pre+Linear & 38400 & 64 & 43.87 $\pm$ 0.69 & 63.01 $\pm$ 0.71 & 75.46 $\pm$ 0.58 & 80.17 $\pm$ 0.51 \\
        \bottomrule
        \end{tabular}
        \begin{tablenotes}\footnotesize
            \item[*] Trained for 100 epochs instead of the default 15 epochs for ProtoTune, since training a classifier on top of a fixed randomly initialized network is expected to require more fine-tuning than starting from a pre-trained network.
        \end{tablenotes}
        \end{threeparttable}
        \end{adjustbox}
    \label{tab:results_n_classes_or_images_full}
\end{table}

\begin{table}
    \centering
    \caption{Accuracy (\%) on $N$-way $K$-shot classification tasks on Mini-ImageNet for methods trained on the CUB training set (5885 images) with a Conv-4 architecture. All results are reported with 95\% confidence intervals over 600 randomly generated test episodes. Results style: \textbf{best} and \underline{second best}.}
    \begin{tabular}{l l |c c c c}
        \toprule
        {\bf Training} & {\bf Testing} & {\bf (5,1)} & {\bf (5,5)} & \bf{(5,20)} & \bf{(5,50)}\\ 
        \midrule
        ProtoCLR & ProtoNet & \underline{34.56} $\pm$ 0.61 & \textbf{52.76} $\pm$ 0.63 & \underline{62.76} $\pm$ 0.59 & \underline{66.01} $\pm$ 0.55 \\
        ProtoCLR & ProtoTransfer & \textbf{35.37} $\pm$ 0.63 & \underline{52.38} $\pm$ 0.66 & \textbf{63.82} $\pm$ 0.59 & \textbf{68.95} $\pm$ 0.57 \\
        Pre(training) & Linear & 33.10 $\pm$ 0.60 & 47.01 $\pm$ 0.65 & 59.94 $\pm$ 0.62 & 65.75 $\pm$ 0.63 \\
        \bottomrule
        \end{tabular}
    \label{tab:results_cub2mini_full}
\end{table}

\end{document}